\documentclass{article}
\PassOptionsToPackage{numbers}{natbib}

\usepackage{amssymb, amsmath, amsthm, mymacros, graphicx, tikz, appendix}
\usepackage{float}
\usepackage{booktabs}
\usepackage[stable]{footmisc}
\usepackage[numbers]{natbib}
\usepackage[bookmarks=true]{hyperref}
\usepackage{multirow}
\usepackage{pbox}
\usepackage{hhline}
\usepackage{subcaption}
\usepackage[font=scriptsize,labelfont=bf]{caption}
\usepackage[margin=1.25in]{geometry}
\newcolumntype{L}{>{\centering\arraybackslash}m{2cm}}
\newcommand{\SO}{\mathrm{SO}}
\newcommand{\LSTM}{\mathrm{LSTM}}

\begin{document}
\title{Lie Access Neural Turing Machine}
\author{Greg Yang$^{a}$
    \thanks{email: gyang@college.harvard.edu}\\
	$^{a}${\em{Harvard University}}    
}

\makeatletter
\let\theauthor\@author
\let\thetitle\@title
\makeatother
\maketitle
 
\begin{abstract}
Following the recent trend in explicit neural memory structures, we present a new design of an external memory, wherein memories are stored in an Euclidean key space $\R^n$.
An LSTM controller performs read and write via specialized read and write heads.
It can move a head by either providing a new address in the key space (aka random access) or moving from its previous position via a Lie group action (aka Lie access).
In this way, the ``L'' and ``R'' instructions of a traditional Turing Machine are generalized to arbitrary elements of a fixed Lie group action.
For this reason, we name this new model the Lie Access Neural Turing Machine, or LANTM.

We tested two different configurations of LANTM against an LSTM baseline in several basic experiments.
We found the right configuration of LANTM to outperform the baseline in all of our experiments.
In particular, we trained LANTM on addition of $k$-digit numbers for $2 \le k \le 16$, but it was able to generalize almost perfectly to $17 \le k \le 32$, all with the number of parameters 2 orders of magnitude below the LSTM baseline.
\end{abstract}
\section{Introduction}

Recurrent neural networks (RNNs) are powerful devices that, unlike conventional neural networks, are able to keep state across time.
They achieved great results in diverse fields like machine translation \cite{sutskever_sequence_2014, cho_learning_2014, bahdanau_neural_2014}, speech recognition \cite{graves_speech_2013, cho_describing_2015}, image captioning \cite{mao_deep_2014, karpathy_deep_2014, vinyals_show_2014}, and many others.
However, despite such advances, traditional RNNs still have trouble maintaining memory for long periods of time, presenting an obstacle to attaining human-like general intelligence.

Following the pioneering work of Graves et al.~\cite{graves_neural_2014} and Weston et al.~\cite{weston_memory_2014}, researchers have studied many variations of external memories equipped to RNNs or explicit memory structures which ameliorate the problem discussed above and obtained great results in applications like question answering \cite{weston_memory_2014, sukhbaatar_end--end_2015, kumar_ask_2015}, algorithm learning \cite{graves_neural_2014, kalchbrenner_grid_2015, kaiser_neural_2015, kurach_neural_2015, zaremba_reinforcement_2015, grefenstette_learning_2015}, machine translation \cite{kalchbrenner_grid_2015}, and others.
In this paper we propose a new variation of external memory.
%

In a conventional RAM used in personal computers, memory is stored at integer addresses, and access is either random or sequential.
Here we replace the integers with $\R^n$, and to retrieve memory, the controller can either issue a brand new address or ``drag'' the previous address in some chosen ``direction'' (formally, apply a Lie group action to the previous address).
The former is the analog of random access, and the latter is the analog of sequential access.
We call the latter ``Lie access,'' with the meaning parametrized by a Lie group $G$ which specifies how this ``dragging'' is to be done.
We call a model built around this concept of ``Lie access'' a Lie Access Neural Turing Machine, or LANTM.
We give two specific implementations in section \ref{LieAccessMemory} and explore them in section \ref{Experiments} with several experiments.
While we will refer to these implementations also as LANTMs, we want to stress they are certainly not the only ways of instantiating the ``Lie access'' concept.

\section{Background}
\subsection{Lie groups}
We assume the reader has a basic knowledge of groups and group actions and the passing notion that Lie groups are just groups with ``differentiable'' operations.
Such a background should enable one to understand the rest of this paper other than section \ref{generalization}.
We defer readers who need slightly more exposition on these topics to Appendix \ref{Liegroups}.

\subsection{Recurrent Neural Networks}

Unlike the conventional feedforward neural network, a recurrent neural network (RNN) has self-connections.
Mathematically, an RNN is a function $\rho: X \times H \to Y \times H$, where $X$ is the input space, $Y$ the output space, and $H$ the space of internal states.
On input $(\p{x}1, \ldots, \p{x}T) \in X^T$ and with initial state $\p{h}0 \in H$, the RNN transitions into states $\p{h}1, \ldots, \p{h}T$ (internally) and returns a sequence $(\p{y}1, \ldots, \p{y}T)$ (externally) defined recursively by
\begin{align*}
(\p{y}t, \p{h}t) & = R(\p{x}t, \p{h}{t-1}).
\end{align*}

%

\begin{figure}[t]
\centering
\includegraphics[scale=1]{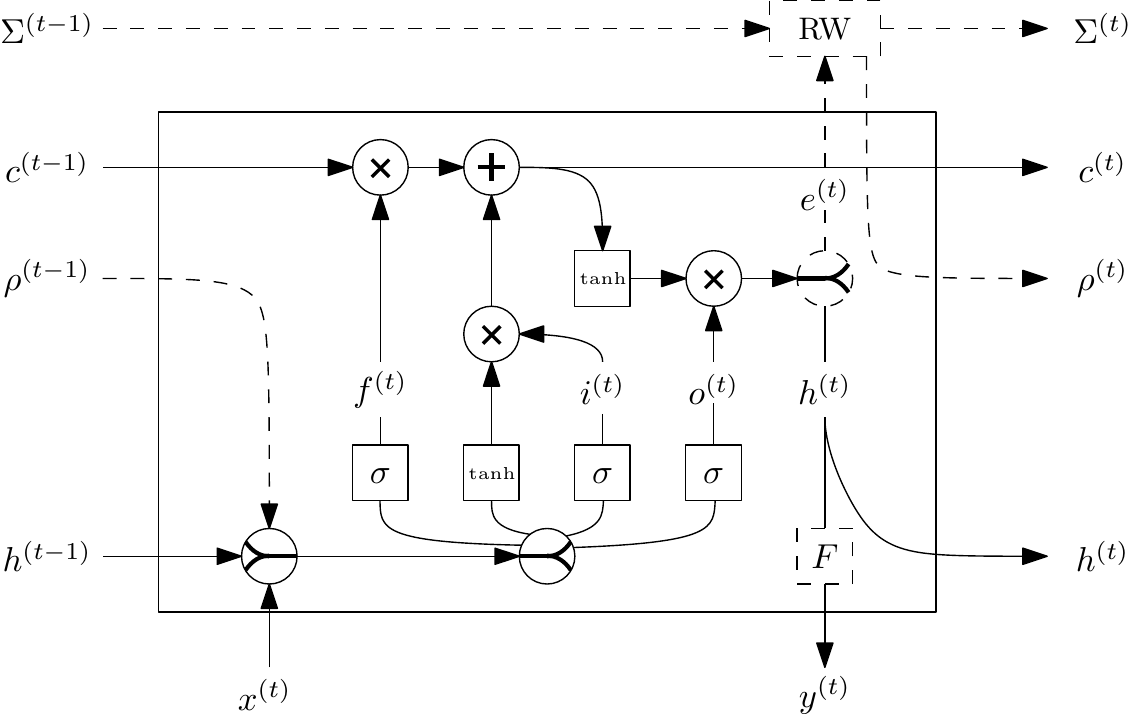}
\caption{LSTM schematics, with and without external memory.
A plain LSTM is illustrated by the undashed part of the diagram.
LSTM as a controller of an external memory is illustrated by including the dashed parts.
The $\succ$ gate indicates concatenating inputs and applying a linear transformation given by the weights of the network.
The $\prec$ gate indicates the splitting of a vector.
$F$ is any processing of $\p{h}t$ to produce the final output $\p{y}t$, e.g. a softmax to produce a distribution over vocabulary.
}
\label{LSTM}
\end{figure}

In this work, we use a particular variant of RNN called the Long Short Term Memory (LSTM) \cite{hochreiter_long_1997}.
LSTM's hidden state consists of two variables $(\p{c}t, \p{h}t)$, where $\p{h}t$ is also the output to the external world (i.e. it fills the role of $\p{y}t$ in the above description).
The $\p{c}t$ is the ``memory'' of the machine, designed to be maintained for a long time when necessary.
There are many variants of LSTM. 
In this paper we define the function $\mathrm{LSTM}: (\p x t, \p c {t-1}, \p h {t-1}) \mapsto (\p y t, \p c t, \p h t)$ as follows:
\begin{align*}
\p{i}t & := \sigma(W_{xi} \p{x}t + W_{hi}\p{h}{t-1} + b_i)\\
\p{f}t & := \sigma(W_{xf}\p{x}t + W_{hf}\p{h}{t-1}  + b_f)\\
\p{c}t & := \p{f}t \p{c}{t-1} + \p{i}t\tanh(W_{xc}\p{x}t + W_{hc}\p{h}{t-1} + b_c)\\
\p{o}t & := \sigma(W_{xo}\p{x}t + W_{ho}\p{h}{t-1} + b_o)\\
\p{h}t & := \p{o}t \tanh(\p{c}t)\\
\p y t & := \p h t
\end{align*}
where $\sigma$ is the logistic function. $\p{i}t, \p{f}t, \p{o}t$ are called the input, forget, and output gates, respectively, which modulate multiplicatively different quantities in the computation.
The weights $W_{\cdot \cdot}$ are trainable through backpropagation through time (BPTT) \cite{werbos_backpropagation_1990}.
The undashed parts of figure \ref{LSTM} show a schematic of the equations above.

In models with external memories, LSTM often serves as the controller \cite{graves_neural_2014,grefenstette_learning_2015,zaremba_reinforcement_2015}.
This means that 1) the entire system carries state over time from both the LSTM and the external memory, 2) the LSTM controller collects reading from and computes additional instructions to the external memory, and 3) the LSTM possibly performs extra processing $F$ to return the desired output at each time point.
The dashed parts of figure \ref{LSTM} demonstrate a typical such arrangement, in which $\p \Sigma t$ represents the state of the memory, $\p \rho t$ represents the reading from the memory, $\mathrm{RW}$ represents a subroutine used for reading from and writing to the memory.
The entire system is now described by the recurrence
$\mathrm{TM}: (\p x t, \p \Sigma {t-1}, \p c {t-1}, \p \rho {t-1}, \p h {t-1}) \mapsto
(\p y t, \p \Sigma t, \p c t, \p \rho t, \p h t)$ defined by
\begin{align*}
(\p e t \oplus \p h t, \p c t, \p e t \oplus \p h t ) &:= \LSTM(\p x t, \p c {t-1}, \p \rho {t-1} \oplus \p h {t-1}) \\
\p y t &:= F(\p h t)\\
(\p \Sigma t, \p \rho t) &:= \mathrm{RW}(\p \Sigma {t-1}, \p e t),
\end{align*}
where $\p e t$ is a set of instructions to read from and write to the memory, as illustrated in figure \ref{LSTM}.
$F$ is usually a softmax layer that produces a distribution over all possible symbols in a language task such as those explored in this paper, and this is indeed the case with LANTM.
In the next section, we show how LANTM implements $\mathrm{RW}$. 
\section{Lie Access Memory}
\label{LieAccessMemory}

The Lie Access Neural Turing Machine (LANTM) is inspired by the external memory architecture of Neural Turing Machine (NTM): a neural network controller reads from and writes to a memory structure via specially designed, differentiable functions called ``heads''.
The heads themselves do not have any trainable parameters, so the only learning done is by the controller, and the entire network can be trained by gradient descent.

In a LANTM, the memory structure is a dictionary, with keys in an Euclidean space $\R^n$ for a fixed $n$, called the {\it key space} or {\it address space}; and with values (called {\it memory vectors}) in another Euclidean space $\R^m$ for a fixed $m$ ($m$ is called the {\it memory width}).
At time step $t$, each read head converts instructions from the controller to a read address $\p k t _r \in
\R^n$ that retrieves a reading $\p \rho t$ from the memory by a weighted inverse squared law, to be elaborated below.
Each write head converts instructions from the controller to a new memory vector $\p m t \in \R^m$ and a new address $\p k t _w \in \R^n$, along with a scalar $\p s t \in [0, 1]$, called the {\it memory strength} of the vector.
Such a triple $(\p k t_w, \p m t, \p s t)$ is essentially appended to the memory.

The most important hyperparameter of a LANTM is its choice of Lie group $G$ that acts on $\R^n$.
At time $t+1$, the controller may emit new addresses for each head (random access) or issue Lie actions $g \in G$ that change the old addresses (Lie access).
One may imagine the key space to be a piece of paper, and the read and write heads to be stones placed on this paper.
The controller is a hand that moves the stones from turn to turn.
Sometimes it may lift a stone up and place it somewhere completely unrelated to its original position (random access); other times it may drag a stone along a chosen direction (Lie access).
Thus Lie access generalizes sequential access in a conventional memory array to a continuous setting.

In the design discussed in this paper, there is no explicit erasure.
However, the machine can theoretically store the exact negation of a memory vector at the same location to cancel out that memory, albeit the required precision to do so would probably be overwhelming.

What follows are details of the overview given above.


\subsection{Read}

\begin{figure}[t]
\centering
\includegraphics[scale=.9]{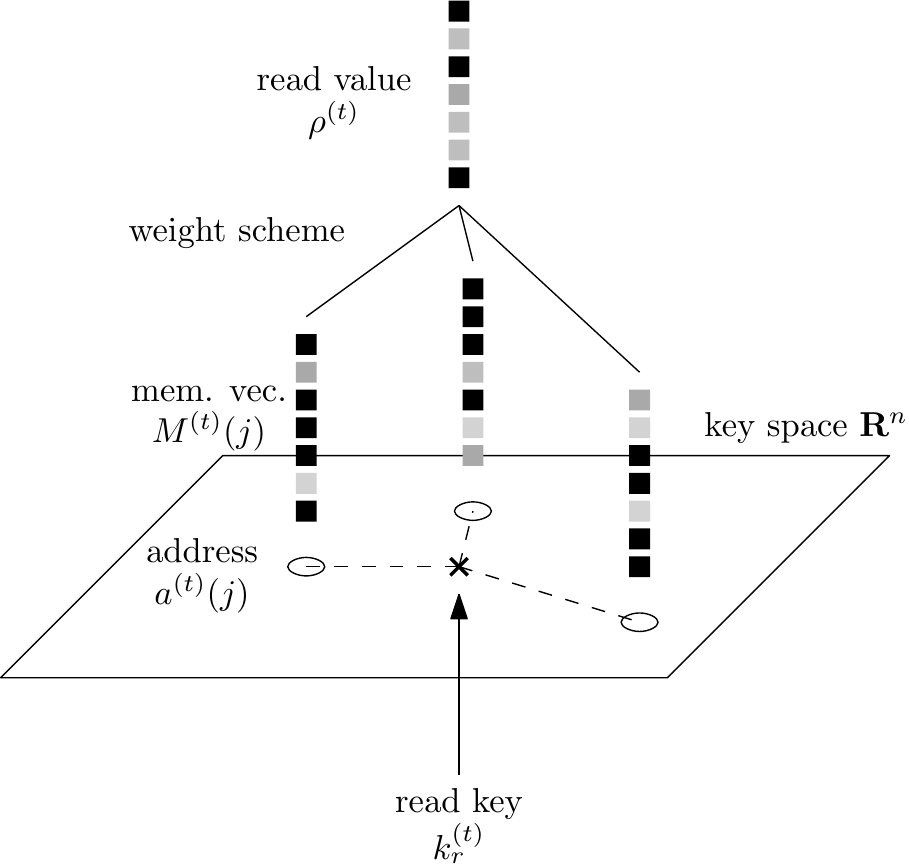}
\caption{Retrieval of value from memory via a key. Weightings with unit sum are assigned to different memories depending on the distances from the addresses to the read key. The weighted arithmetic mean is emitted as the final read value.
Both InvNorm and SoftMax schemes follow this method, but each with a different way of computing the weightings.
In particular, the SoftMax scheme requires another input, the temperature $\p T t$.}
\label{retrieval}
\end{figure}

Let $\p M t$ denote the set of memory vectors stored in the key space by time
$t$.
We choose a canonical ordering on this set, for example by time added, and write
$\p M t(i)$ for the $i$th vector in this order.
Denote by $\p a t(i)$ the corresponding addresses of $\p M t(i)$ and by $\p S t(i)$ the corresponding memory strength of $\p M t(i)$.
In this section we introduce two {\it weight schemes} for retrieving a value from the memory via an address.
The main idea of both is summarized by figure \ref{retrieval}.

The read key $\p k t _r$ produces weightings $\p w t _r(i)$ over all memory
vectors $\p M t (i)$, each with address $\p a t (i)$, by normalizing their
inverse squared distances and multiplying by their strengths $\p S t (i)$:
$$\p w t _r(i) := \f{\p S t(i)\|\p k t _r - \p a t (i)\|^{-2}}
					{\sum_j \|\p k t _r - \p a t (j)\|^{-2}}$$
with the convention that it takes the limit value when
$\p k t _r \to \p a t (i)$
for some $i$.
\footnote{
In practice, as the formula for $\p w t _r$ can induce numerical instability as $\p k t _r \to \p a t (i)$ for some $i$, we adjust the formula with a small $\epsilon$, e.g. $10^{-9}$, so that
$$\p w t _r(i) := \f{\p S t(i)(\|\p k t _r - \p a t (i)\|^{2} + \epsilon)^{-2}}
				{\sum_j (\|\p k t _r - \p a t (j)\|^{2} + \epsilon)^{-2}}.$$
}

The reading is then defined as
$$\p \rho t := \sum_j \p w t_r (j) \p M t (j)$$


We call this method of converting a read key to a set of weighting via a polynomial law {\it InvNormalize}, or {\it InvNorm} for short, 
in contrast with the use of exponential law in the case of SoftMax weight scheme, which computes the weights $\p w t _r(i)$ as
$$\p w t _r(i) := \f{\p S t(i) \exp(-\|\p k t _r - \p a t (i)\|^{2}/\p T t)}
		{\sum_j \exp(-\|\p k t _r - \p a t (j)\|^{2}/\p T t)}
$$
where $\p T t$ is a {\it temperature} emitted by the controller at time $t$ that represent the certainty of its reading.
The higher $\p T t$ is, the more $\p w t _r$ tends to be uniform.

Given the ubiquity of SoftMax in the machine learning literature, one may consider it a natural choice for the weight scheme.
But as will be seen in the experiments, InvNorm is crucial in making the Euclidean space work as an address space.

\subsection{Write}

There is no extra ingredient to writing other than adding the produced memory vector $\p m t$, its strength $\p s t$, and its address $\p k t_w$ to the collection of memory vectors, strengths, and addresses.
To ensure that memory selection by weighted average works well, we squash the values of $\p m t$ to $[-1, 1]$ by $\tanh$, but squashing by the logistic sigmoid function is also conceivable.
Without such squashing, a memory vector $\p M t (i)$ with large values can dominate the output of a weight method despite having low weight $\p w t_r (i)$. 

\subsection{Addressing procedure}

Here we describe how the keys $\p k t_r$ and $\p k t _w$ are produced. 
The procedure is the same for both read and write keys, so we assume that we are to compute a single key $\p k t$.
We describe the abstraction of the process over a fixed Lie group $G$ acting smoothly on the key space $\R^n$.

\begin{figure}[t]
\centering
\includegraphics[scale=.9]{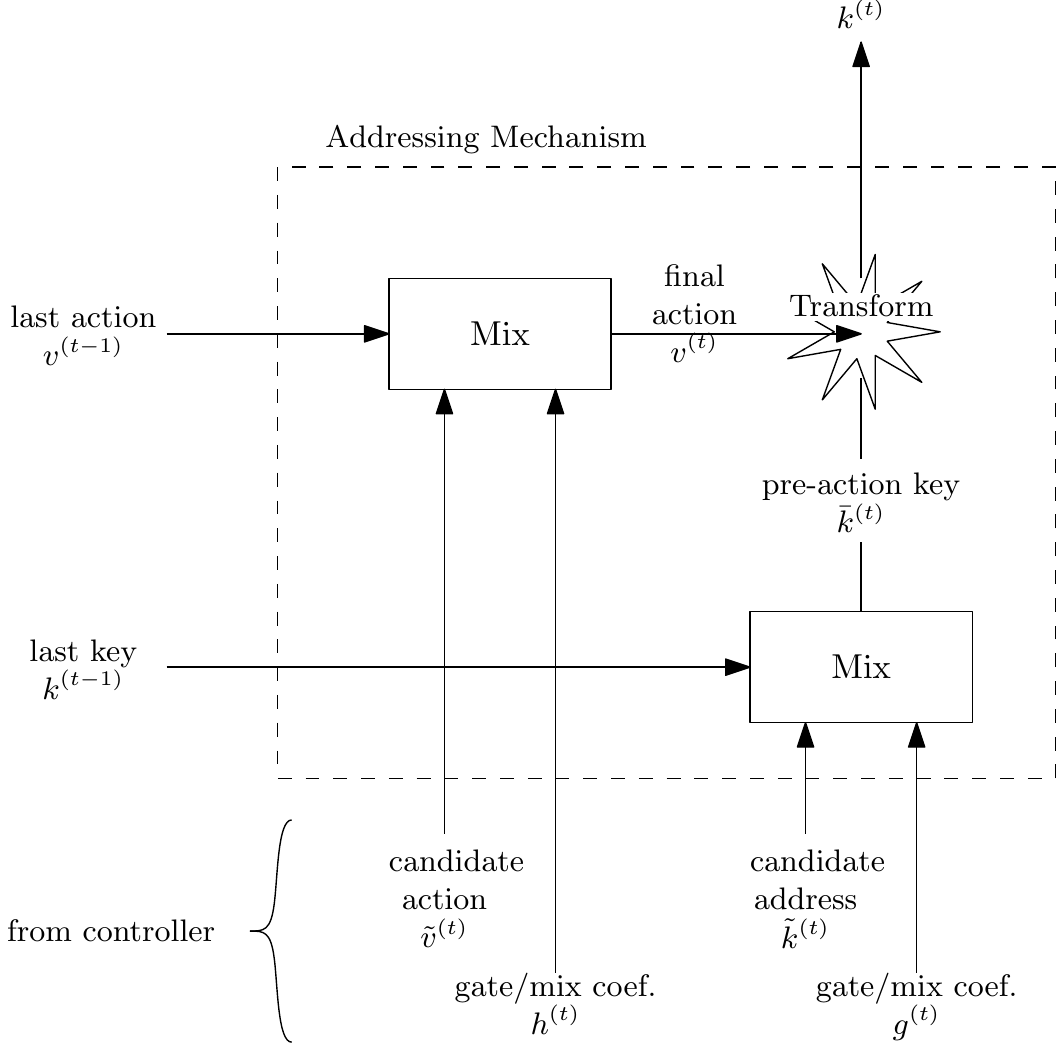}
\caption{addressing mechanism.\label{address_mech}}
\end{figure}

The controller emits 3 things: a {\it candidate key} $\p{\tilde k} t \in \R^n$, a {\it mixing coefficient}, or {\it gate}, $\p g t \in [0, 1]$ (via the sigmoid function), and an action $\p v t \in G$ that we also call {\it step}.
The gate $g$ mixes the previous key $\p k {t-1}$ with the candidate key to produce a {\it pre-action key} $\p {\bar k} t$, which is transformed by $\p v t$ to produce the final key $\p k t$: (here $\cdot$ denotes group action)
\begin{align*}
\p {\bar k} t &:= \p g t \p {\tilde k} t + (1 - \p g t) \p k {t-1}\\
\p k t & := \p v t \cdot \p{\bar k} t.
\end{align*}
Figure (\ref{address_mech}) summarizes the addressing procedure.

In our experiments, the Lie group is $\R^2$ acting additively on $\R^2$.
This means that the controller outputs 2 numbers $a = \p a t, b = \p b t$, so that $v = (a, b)$ acts upon a key $k = (x, y)$ by
$$v\cdot k = (x, y) + (a, b) = (x + a, y + b).$$
Section \ref{ExampleReps} in the Appendix gives example implementations for the scaling rotation $\R^* \times \SO(2)$ and the rotation groups $\SO(2)$ acting on $\R^2$.

\subsection{Interpolation of Lie action}

For readers unfamiliar with the Lie group examples mentioned below, we recommend a visit to section \ref{ExampleReps} in the Appendix.

For groups like $(\R^n, +)$, there is a well-defined convex interpolation between two elements that stays in the group.
For some others like $\R^* \times \SO(2)$, the straight-line interpolation $tv + (1 - t)w$ for $t \in [0, 1]$, $v, w \in G$ sometimes produce elements outside the group (in this case sometimes the elements cancel out and get 0), but does so with probability zero in a suitable sense.

Then, as for keys, we can let the controller output a candidate action $\p{\tilde v} t\in G$ and a mixing coefficient $\p h t$ to smoothly mix with the previous action $\p {v} {t-1}$ to produce a final action 
$$\p v t := \p h t \p {\tilde v} t + (1 - \p h t) \p v {t-1}.$$

This allows the controller to ``move in a straight line within the group of actions'' by merely left saturating (i.e. squash to 0) the gates $\p g t$ and $\p h t$ for all $t$, so that $\p v 1 = \p v 2 = \p v 3 = \cdots$.
Of course, the ``straight line'' can be actually curved depending on the group.
For example, when $G = \R^* \times \SO(2)$, a typical ``straight line'' will be a spiral tending exponentially toward the origin or growing exponentially unbounded.

Even if a group doesn't have a natural straight-line interpolation, there may be another way to mix two actions.
In the case of $G = \SO(2) \cong S^1$, we can just project a straight-line interpolation onto the circle (barring a measure zero chance of intepolating into $(0, 0) \in \R^2$).
\footnote{
There is, in fact, a canonical way to interpolate the most common Lie groups, including all of the groups mentioned above, based on the exponential map and the Baker-Campbell-Hausdorff formula \cite{lee_introduction_2012}, but the details are outside the scope of this paper and the computational cost, while acceptable in control theory settings, is too hefty for us.
Interested readers are referred to \cite{shingel_interpolation_2009} and \cite{marthinsen_interpolation_1999}.
}

\begin{figure}[t]
\centering
\includegraphics[scale=.7]{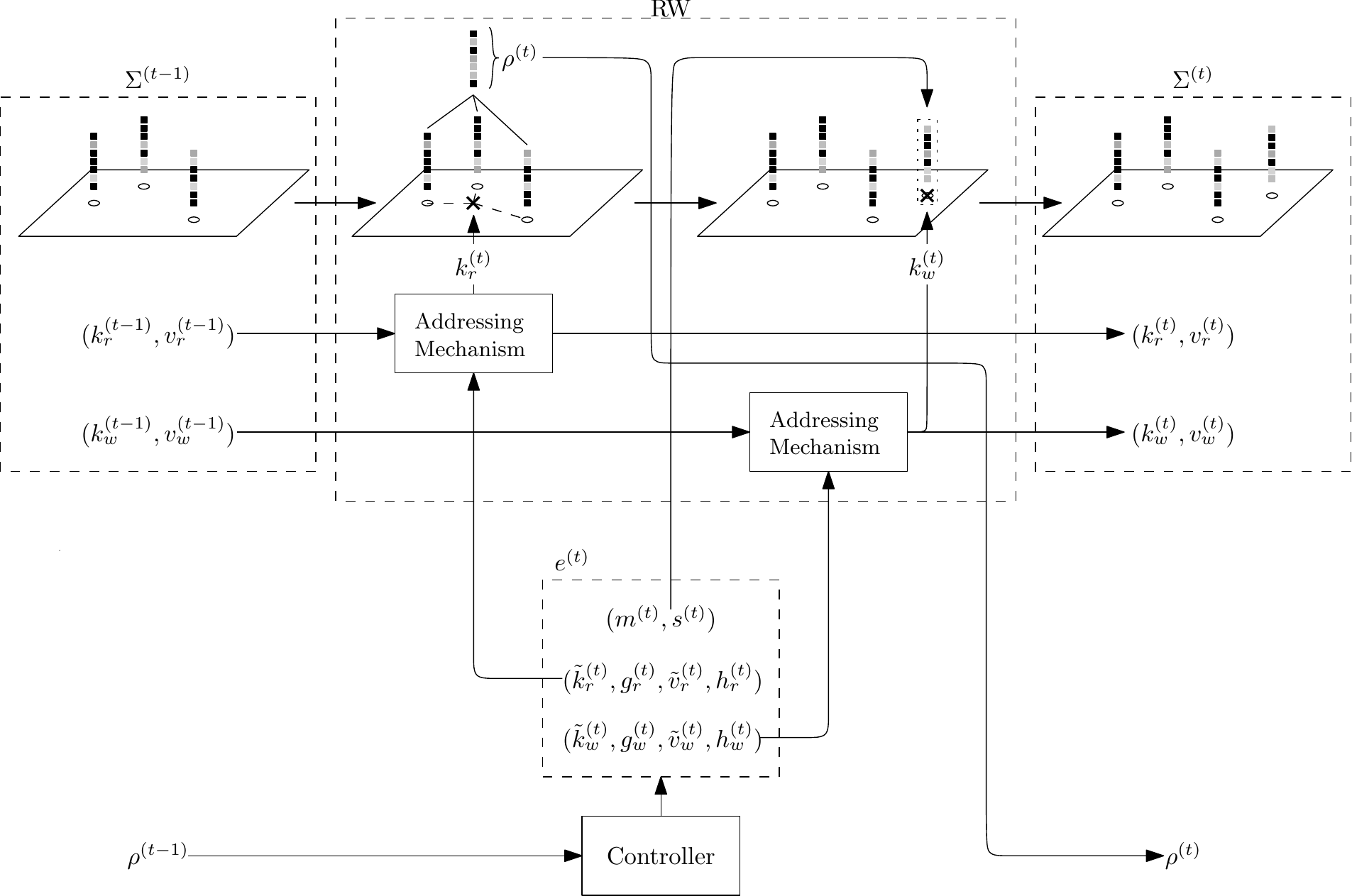}
\caption{Summary of controller interaction with external memories.
The dashed boxes correspond to dashed parts in figure \ref{LSTM}.
Note that all input, output and the states of the LSTM other than $\p \rho t$ have been omitted. 
\label{controller_mem}}
\end{figure}

The final addressing mechanism is shown in figure \ref{address_mech}.
All together, the interaction of the controller with the external memory is shown in figure \ref{controller_mem}.

\section{Experiments}
\label{Experiments}
In our experiments, the Lie group for both types of LANTM is the translation group $\R^2$ acting on $\R^2$ \footnote{We early on experimented with the scaling rotation group $\R^* \times \SO(2)$, which produced acceptable results when input lengths were small but encountered numerical problems when input lengths were large due to exponentiating scale.}, and we used Lie action interpolation as specified above. 
We outline the most important experimental setup in the main text below but defer other details to the Appendix section \ref{exp-details}. 

\subsection{permutation and arithmetic tasks}
We tested the two variations of LANTM along with a baseline LSTM in an encoder-decoder setup (cf. \cite{sutskever_sequence_2014}) on the copy, reverse, and bigram flip tasks as done in \cite{grefenstette_learning_2015}, as well as the double and addition tasks designed in a similar vein.
Table \ref{permutation_inputoutput} shows input/output templates for each permutation task.

\begin{table}[ht]
\centering
\caption{Input/output templates for permutation tasks}
\scriptsize {
\begin{tabular}{lllll}
\toprule
task       & input & output\\ \midrule
copy       & $a_1 a_2 a_3 \cdots a_k$ & $a_1 a_2 a_3 \cdots a_k$\\
reverse    & $a_1 a_2 a_3 \cdots a_k$ & $a_k a_{k-1} a_{k-2} \cdots a_1$\\
bigramFlip & $a_1 a_2 a_3 a_4 \cdots a_{2k-1} a_{2k}$ & $a_2 a_1 a_4 a_3 \cdots a_{2k} a_{2k-1}$\\
\bottomrule      
\end{tabular}
}
\label{permutation_inputoutput}
\end{table}

Each arithmetic tasks have all numbers, input or output, formatted with the least significant digits \emph{on the left} and with zero padding.
The double task takes an integer $x \in [0, 10^k)$ padded to $k$ digits and outputs $2x$ in $k+1$ digits, zero padded to $k+1$ digits. 
The addition task takes two integers $x, y \in [0, 10^k)$ padded to $k$ digits and \emph{interleaved}, forming a length $2k$ input sequence and outputs $x+y$ zero padded to $k+1$ digits.
Table \ref{arithmetic_inputoutput} show example input/outputs for each task with $k=3$.

\begin{table}[ht]

\parbox{.4\linewidth}{
\centering
\caption{Input/output examples for arithmetic tasks}
\scriptsize {
\begin{tabular}{lllll}
\toprule
task       & input & output & explanation\\ \midrule
double       & $928$  & $8561$ & $2 * 829 = 1658$\\
addition    & $439204$ & $7150$ & $423 + 94 = 517$\\
\bottomrule      
\end{tabular}
}
\label{arithmetic_inputoutput}
}
\parbox{.59\linewidth}{

\centering
\caption{Input sequence lengths or operand digits for each task.}
\scriptsize {
\begin{tabular}{lllll}
\toprule
task       & min train & max train & min test & max test \\ \midrule
copy       & 2         & 64        & 65       & 128      \\
reverse    & 2         & 64        & 65       & 128      \\
bigramFlip & 2         & 32        & 33       & 64       \\
double     & 2         & 40        & 41       & 80       \\
addition   & 2         & 16        & 17       & 32		\\
\bottomrule      
\end{tabular}
}
\label{input_sizes}}
\end{table}

\begin{figure}[t]
\centering
\includegraphics[scale=.8]{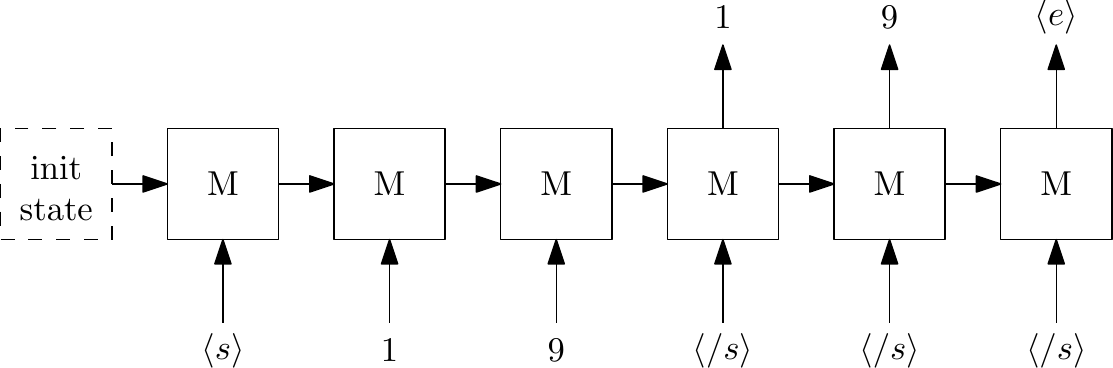} 
\caption{example in/out schematic. \label{example_inout}}
\end{figure}

The machines are first fed a learnable initial state and then provided with the input sequence, flanked by a start-of-input (SOI) symbol $\la s \ra$ and a repetition of an end-of-input (EOI) symbol $\la /s\ra$.
The machines are to output the correct sequence during the {\it response phase}, which starts when they receive the first $\la /s \ra$.
The repetition of $\la /s \ra$ effectively means that the correct symbols are not shown to the machines during answering, i.e. we do not use teacher forcing.
The machine also must correctly emit an end-of-output (EOO) symbol $\la e \ra$ to terminate their answers.
Figure (\ref{example_inout}) is an example of inputs and correct outputs during a copy task.

As usual, prediction is performed via argmax but training is done by minimizing negative log likelihood.
To evaluate the performance of the models, we compute the fraction of characters correctly predicted and the fraction of all answers completely correctly predicted, respectively called ``fine score'' and ``coarse score'' following \cite{grefenstette_learning_2015}.


{\bf Task parameters and hyperparameters.} We trained the models on the above tasks for input sizes summarized by table \ref{input_sizes}.
For all tasks, the LANTM has a single-layer, 50-cell or 100-cell LSTM controller.
The memory width (i.e. the size of each memory vector) is 20.
For all tasks, the LSTM baseline has 4 layers, each with 256 cells.
In the Appendix, the exact parameters for each model in each task are listed in table \ref{param_sizes}, and other experimental details are given in section \ref{exp-details}.
Notice that the LSTM has 2 orders of magnitude more parameters than the LANTM models.

{\bf Results.} LANTM-InvNorm was able to master all tasks and generalize nearly perfectly to 2x the training sizes, as shown in table \ref{exp-results}.
LANTM-SoftMax did as well on the copy and double tasks but failed at all the others, having performed worse than the LSTM baseline.
The baseline itself learned tasks with smaller training input sizes (bigramFlip, double, addition) almost flawlessly, but generalization to 2x training size was inadequate on all tasks, with coarse score not exceeding 6\%.

\begin{table}[ht]
\caption{Permutation and arithemetic task results. ``$n$ x'' indicates tested sequence length compared to the trained length.
All values are rounded to the nearest integer percent.}
\centering
\tiny {
\begin{tabular}{llllll}
\toprule
task                        & model         & 1x coarse & 1x fine & 2x coarse & 2x fine \\ 
\midrule
\multirow{3}{*}{copy}       & LANTM-InvNorm & 100\%      & 100\%    & 100\%      & 100\%    \\
                            & LANTM-SoftMax & 100\%      & 100\%    & 99\%       & 100\%    \\
                            & LSTM          & 58\%       & 97\%     & 0\%        & 52\%     \\ \midrule
\multirow{3}{*}{reverse}    & LANTM-InvNorm & 100\%      & 100\%    & 100\%      & 100\%    \\
                            & LANTM-SoftMax & 1\%        & 12\%     & 0\%        & 4\%      \\
                            & LSTM          & 65\%       & 95\%     & 0\%        & 44\%     \\ \midrule
\multirow{3}{*}{bigramFlip} & LANTM-InvNorm & 100\%      & 100\%    & 99\%       & 100\%    \\
                            & LANTM-SoftMax & 12\%       & 40\%     & 0\%        & 10\%     \\
                            & LSTM          & 98\%       & 100\%    & 4\%        & 58\%     \\ \midrule
\multirow{3}{*}{double}     & LANTM-InvNorm & 100\%      & 100\%    & 100\%      & 100\%    \\
                            & LANTM-SoftMax & 100\%      & 100\%    & 100\%      & 100\%    \\
                            & LSTM          & 98\%       & 100\%    & 2\%        & 60\% \\  \midrule
\multirow{3}{*}{addition}   & LANTM-InvNorm & 100\%      & 100\%    & 99\%       & 100\%    \\
                            & LANTM-SoftMax & 17\%       & 61\%     & 0\%        & 29\%     \\
                            & LSTM          & 97\%       & 100\%    & 6\%        & 64\%  
\\ \bottomrule
\end{tabular}
}
\label{exp-results}
\end{table}

We tested the learned InvNorm model on larger, arbitrarily selected input sizes.
The results are summarized by table \ref{explore-gen}.
On permutation tasks, it generalized quite well when challenged by 4 times the training size, able to get more than 90\% of test problems correct.
On the double task, its extrapolation performance was similar, with 86\% coarse score on 4x training size.
Notice that LANTM-InvNorm on several of the tasks (8x bigramFlip, 8x double, 4x addition) achieved high fine scores when extrapolating to large input sizes despite having low coarse scores.
This suggests that the extrapolation errors systematically occur at the end of each output on those tasks.

\begin{table}[ht]
\caption{Exploring the Generalizability of LANTM-InvNorm. 
}
\centering
\tiny{
\begin{tabular}{lllllll}
\toprule
task       & 4x coarse & 4x fine & 5x coarse & 5x fine & 8x coarse & 8x fine \\ \midrule
copy       & 100\%      & 100\%    & 91\%       & 100\%    &            &          \\
reverse    & 91\%       & 98\%     & 12\%       & 65\%     &            &          \\
bigramFlip & 96\%       & 100\%    &            &          & 12\%       & 96\%     \\
double     & 86\%       & 99\%     &            &          & 21\%       & 90\%     \\
addition   & 2\%        & 95\%     &            &          &            &
\\ \bottomrule         
\end{tabular}
}
\label{explore-gen}
\end{table}

We have created videos of the read and write locations of LANTM-InvNorm and LANTM-SoftMax while learning each of the 5 tasks, tracking their progress over time.
They are available in the Supplementary Materials, with details explained in appendix \ref{gifs}.
In appendix \ref{close_anal}, we look at the behaviors of trained LANTM-InvNorm through their read and write locations, gate values, and example input/output to analyze what exactly they learned and where their extrapolation errors come from when challenged by extreme input lengths.

\subsection{Python programs}

The above problem setting is highly structured and favors the design of LANTM.
In this task we trained the models on generated python programs, following \cite{zaremba_learning_2014}, that is more natural.
The dataset comprises of 6 types of programs of integers: addition/subtraction, identity,  multiplication with one small operand, small for loops, variable substitution, and ternary ``$a$ if $b$ else $c$'' statements, as illustrated in table \ref{python_input_output}.

The models are required to read the input program, which terminates with a ``print'' statement, and output the correct integer response, \emph{in reverse sequence}, without being fed the correct answer (same as in our last experiment, but different from \cite{zaremba_learning_2014}, which used teacher forcing).
We performed curriculum learning, using the ``mixed'' strategy of \cite{zaremba_learning_2014}, starting from 2 digits operands up to 4 digits operands.
We evaluated the models on their coarse and fine scores on randomly sampled 4 digit programs.
Training was done by RMSProp with learning rate 0.002, which was multiplied by 0.8 whenever the validation accuracy became lower than the highest of the last four.

Here the LSTM baseline is a single layer of 128 cells, and the LANTM models also have controllers who have the same size.
In addition, each LANTM model has memory size 128.

The results are summarized by table \ref{L2Xresults}.
We noted that the small loop programs were the most difficult program type, for which all models predicted less than half of the characters correctly,
so we trained them in a separate experiment only on small loop programs.
The results are given in table \ref{smallloopresults}

\begin{table}[ht]
\parbox{.45\linewidth}{
\centering
\caption{Results for learning python programs}
\tiny {
\begin{tabular}{lllll}
\toprule
model      & coarse    & fine \\ \midrule
LSTM       & 35\%      & 66\% \\ 
LANTM-InvNorm & 39\%         & 74\%\\ 
LANTM-SoftMax & 35\%         & 67\% \\ 
\bottomrule      
\end{tabular}
}
\label{L2Xresults}
}
\parbox{.45\linewidth}{
\centering
\caption{Results for learning small loop programs}
\tiny {
\begin{tabular}{lllll}
\toprule
model      & coarse    & fine \\ \midrule
LSTM       & 0\%      & 51\% \\ 
LANTM-InvNorm & 0\%         & 55\%\\ 
LANTM-SoftMax & 0\%         & 55\%\\
\bottomrule      
\end{tabular}
}
\label{smallloopresults}
}
\end{table}

Here the advantage of LANTM over LSTM is not as dramatic.
The memory access of LANTM were not nearly as orderly and neat as in the previous experiment, but rather erratic looking.
An interactive plot of example read and write locations and other state data of LANTM-InvNorm while learning small loops can be found in the Supplementary Materials.

\subsection{language modelling}

Finally, we tested the models on the Penn treebank corpus.
To train and predict continuously, whenever the external memories of LANTMs were fill up to 100 memory vectors, the oldest 60 vectors were discarded.
As in the last experiment, the LSTM baseline is a single layer of 128 cells, and the LANTM models also have controllers with the same size.
In addition, each LANTM model has memory size 128.
We unrolled BPTT to 20 steps, and trained with Adagrad with learning rate 0.05, which was halved each time the validation perplexity exceeded that of the previous epoch.

\begin{table}[ht]
\centering
\caption{Perplexity for language modelling corpus}
\tiny {
\begin{tabular}{lllll}
\toprule
model      & validation & test\\ \midrule
LSTM       & 130      & 124 \\ 
LANTM-InvNorm & 128         & 123\\ 
LANTM-SoftMax & 134         & 130\\ 
\bottomrule      
\end{tabular}
\label{langmod_results}
}
\end{table}

We observed that LANTM-InvNorm had its read and write locations at two distant clusters, so that its read weights were all diffuse across the entire memory.
This may be due to the repeated application of a (approximately) single Lie action over the long course of training, blowing up the magnitude of keys, which degrades random access, as the typical squashing functions of the controller limits the range of keys it can produce.
This means that, rather than storing useful information at particular locations, the machine stored \emph{deltas} at each time step, so that the whole memory averaged together gave the desired information.
LANTM-SoftMax also exhibited the same behavior, but because high fidelity access only required the read key to be closer to the desired key $k$ much more than to other keys (rather than that its distance to $k$ be absolutely small as with InvNorm), we cannot immediately infer that it also only stored deltas.

\section{Related Works}

Zaremba et al.~\cite{zaremba_learning_2014} taught LSTM to evaluate simple python programs via curriculum learning, which formed the basis of one of our experiments.
Kalchbrenner et al.~\cite{kalchbrenner_grid_2015} arranged LSTM cells in a multidimensional grid to form the \emph{grid long short term memory}, and learned copy and addition tasks as well.
Graves et al.~\cite{graves_neural_2014} created NTM which has inspired much of the design in our work.
Zhang et al.~\cite{zhang_structured_2015} found several tweaks to NTM to improve its convergence and performance.
Grefenstette et al.~\cite{grefenstette_learning_2015} designed smooth versions of stack, queue, and deque as external memories to an LSTM controller.
Their unbounded memory and experimental setups were direct influences on this paper.
Zaremba et al.~\cite{zaremba_reinforcement_2015} used reinforcement learning to absolve the need of the NTM to involve the entire memory during memory retrieval.
Weston et al.~\cite{weston_memory_2014} came upon similar ideas in the \emph{memory network} as the NTM at around the same time, but with less focus on sequence learning and more on question answering tasks (QA).
Sukhbaater et al.~\cite{sukhbaatar_end--end_2015} improved on their results to give a memory network trainable via gradient descent end-to-end and allowing multiple adaptive memory queries (``multiple hops'') which help in complex relational reasoning.
\emph{Dynamic memory network} of Kumar et al.~\cite{kumar_ask_2015} added an episodic memory module similar to the multiple hops feature of Sukhbaatar et al.'s model, but which dynamically chose when to stop accessing memory rather than after a fixed number of times.
They achieved state of art results in several tasks such as QA and sequence modelling.
Danihelka et al.~\cite{danihelka_associative_2016} designed an external memory based on holographic reduced representations, which can store unlimited memory but the larger the size the more noisy the retrieval.
Kaiser et al.~\cite{kaiser_neural_2015} created the \emph{neural GPU} based on convolutional kernels, which learned long multiplication of binary numbers up to 20 bits but were able to generalize to 2000 bits.
Kurach et al.~\cite{kurach_neural_2015} generalized tha random access of conventional RAMs to create the \emph{Neural Random Access Machine}, which learned simple algorithm and was able to generalize to larger lengths, and memory access during inference can be done in constant time.
Neelakantan et al.~\cite{neelakantan_adding_2015} investigated adding gradient noise to training, and found that in many of the models mentioned above, this method improved the performance or allowed a greater percentage of random initializations to converge to the optimum.

\section[Generalization and Theoretical Considerations]{Generalization and Theoretical Considerations
\footnote{This part mentions some advanced mathematical concepts but is not necessary to the understanding of the rest of the paper}
}
\label{generalization}
We want to stress that the model explained \ref{LieAccessMemory} is but one way to implement Lie access memory.
Indeed, the Euclidean key space could be generalized to any Riemannian manifold equipped with a subgroup of its isometry group, as 1) a notion of metric is required in Lie access memory (hence the Riemannian part), and 2) one wants the ability to store and retrieve information in a ``straight line'' which suggests that the Lie action be invariant with respect to the metric (hence the isometry part).

A potentially useful Riemannian manifold other than $\R^n$ is the hyperbolic space, specifically the Poincare disk model \cite{lee_riemannian_1997}.
As seen in the language modelling task, repeated application of Lie action on $\R^n$ may blow up the magnitude of keys, degrading random access.
The Poincare disk model has its points in the (open) unit ball that prevents this problem from occurring.
The other standard Riemannian model, the sphere, is not quite as desirable in this setting, because it ``wraps around'' (i.e. is not acyclic, in homological/homotopic terms), which can confuse gradient descent.

\section{Conclusion}

In this paper we introduced Lie access memory and explored two different implementations in the experiments.
The LANTM model with the InvNorm weight scheme in all tasks performed better than the baseline, and spectacularly so in sequence and addition tasks where it learned to generalize to extraordinary lengths, whereas that with the SoftMax weight scheme failed to outperform the baseline in the reverse, bigramFlip, addition, and language modelling tasks. 
LANTM-InvNorm held its largest advantage over LSTMs in case of long, structured tasks.

The Python program experiment shows that in less structured environments or environments with redundant or useless information, our LANTM designs could not utilize their memory as impressively as in more structure environments.
Thus further work needs to be done toward combining logical reasoning with natural language processing.

We adopted a simple way to turn the episodic nature of our unbounded memory to continuous use, but it was far from perfect.
In the language modelling experiment, the LANTM models did not seem to use the memory in a remarkable way.
Future work should explore different options for adapting Lie access memory to continuous tasks, for example, by bounding the memory or by using the Poincare disk model as the underlying manifold as suggested in section \ref{generalization}.

\newpage
{\scriptsize
\bibliography{lantm}
\bibliographystyle{plain}}
\newpage
\appendix
\renewcommand\thefigure{\thesection.\arabic{figure}}
\setcounter{table}{0}
\renewcommand*{\thetable}{\thesection.\arabic{table}}

\setcounter{figure}{0}
\appendixpage

\section{Experimental details}
\label{exp-details}
\subsection{permutation and arithmetic tasks}

The baselines of our experiments are LSTMs in an encoder-decoder setup as described in \cite{sutskever_sequence_2014}.
We tested 2 variations of LANTM with an InvNorm and a SoftMax address mechanism, along with the LSTM baseline, on the permutation and arithmetic tasks to be described.
The Lie group for both types of LANTM is the translation group $\R^2$ acting on $\R^2$ \footnote{We early on experimented with the scaling rotation group $\R^* \times \SO(2)$, which produced acceptable results when input lengths were small but encountered numerical problems when input lengths were large due to exponentiating scale.}.
For both LANTMs and LSTM, we embed the input vocabulary continuously via a real embedding matrix into an Euclidean space before feeding into the models;
we also pass the outputs through a softmax layer to arrive at probability distributions over the vocabulary set (this is the $F$ box in figure \ref{LSTM}).
As usual, prediction is performed via argmax but training is done by minimizing negative log likelihood.

The machines are first fed a learnable initial state and then provided with the input sequence, flanked by a start-of-input (SOI) symbol $\la s \ra$ and a repetition of an end-of-input (EOI) symbol $\la /s\ra$.
The machines are to output the correct sequence during the {\it response phase}, which starts when they receive the first $\la /s \ra$.
The repetition of $\la /s \ra$ effectively ensures that the correct symbols are not shown to the machines during answering.
The machine also must correctly emit an end-of-output (EOO) symbol $\la e \ra$ to terminate their answers.
The LANTM models are not allowed to write to the memory during the response phase, so that there is more emphasis on collecting the right information during the input phase.
Figure (\ref{example_inout}) is an example of inputs and correct outputs during a copy task.

{\bf Tasks.} Each task has a length parameter $k$.
The permutation tasks include 
\begin{enumerate}
  \item copy
  \begin{align*}
 	\text{input: } & a_1 a_2 a_3 \cdots a_k \\
	\text{output: } & a_1 a_2 a_3 \cdots a_k
  \end{align*} 

  \item reverse
  \begin{align*}
  \text{input: }& a_1 a_2 a_3 \cdots a_k \\ 
  \text{output: }& a_k a_{k-1} a_{k-2} \cdots a_1
  \end{align*}
  \item bigramFlip
  \begin{align*}
  \text{input: }& a_1 a_2 a_3 a_4 \cdots a_{2k-1} a_{2k} \\ 
  \text{output: }& a_2 a_1 a_4 a_3 \cdots a_{2k} a_{2k-1}
  \end{align*}
%
\end{enumerate}

The arithmetic tasks include the following. Note that all numbers, input or output, are formatted with the least significant digits {\bf on the left} and with zero padding. 
\begin{enumerate}
	\item double. Let $x$ be an integer in the range $[0, 10^k]$, with zero padding in front (on the right) to make up $k$ digits.
	\begin{align*}
	\text{input: }& \text{$x$ in base 10, zero padded to $k$ digits} \\
	\text{output: }& \text{$2 x$ in base 10, zero padded to $k+1$ digits}
	\end{align*}
	
	\item addition. Let $x$ and $y$ be integers in the range $[0, 10^k]$, with zero padding in front (on the right) to make up $k$ digits. If they have digits $x_1x_2\cdots x_k$ and $y_1y_2\cdots y_k$, respectively, with the {\it least} significant digits on the left, then
	\begin{align*}
	\text{input: }&x_1 y_1 x_2 y_2 \cdots x_k y_k \\
	\text{output: }&x + y \text{ in base 10, zero padded to $k+1$ digits, with least significant digits on the left}
	\end{align*}
	\\
	In other words, we interleave the inputs. Thus this is a different encoding of the addition problem from previous works like \cite{zaremba_learning_2014} and \cite{jozefowicz_empirical_2015}.
\end{enumerate}

{\bf Task parameters and hyperparameters.} We trained the models on the above tasks for input sizes summarized by table \ref{input_sizes}.
For all tasks, the LANTM has a single-layer, 50-cell or 100-cell LSTM controller.
The Lie group for all LANTMs is the translation group $\R^2$ acting on the key space $\R^2$.
The memory width (i.e. the size of each memory vector) is 20.
For all tasks, the LSTM baseline has 4 layers, each with 256 cells.
The exact setting of parameters for each model in each task is listed in table \ref{param_sizes}.

\begin{minipage}{\linewidth}
\scriptsize{
\begin{center}
\captionof{table}{Parameters in each model.}
\begin{tabular}{llrrrrrr}
\toprule
Model                          & Task       & LSTM size & Vocab & Embed. & Mem. width & LR & \#Param \\ \toprule
\multirow{5}{*}{\pbox{20cm}{LANTM\\InvNorm}} & copy       & 50                   & 128        & 7              & 20           & 0.02          & 26105        \\ \cline{2-8} 
                               & reverse    & 50                   & 128        & 7              & 20           & 0.02          & 26105        \\ \cline{2-8} 
                               & bigramFlip & 100                  & 128        & 7              & 20           & 0.02          & 70155        \\ \cline{2-8} 
                               & addition   & 50                   & 14         & 14             & 20           & 0.01          & 20291        \\ \cline{2-8} 
                               & double     & 50                   & 14         & 7              & 20           & 0.02          & 18695        \\ \hline
\multirow{5}{*}{\pbox{20cm}{LANTM\\SoftMax}} & copy       & 50                   & 128        & 7              & 20           & 0.02          & 26156        \\ \cline{2-8} 
                               & reverse    & 50                   & 128        & 7              & 20           & 0.02          & 26156        \\ \cline{2-8} 
                               & bigramFlip & 100                  & 128        & 10             & 20           & 0.02          & 72123        \\ \cline{2-8} 
                               & addition   & 50                   & 14         & 14             & 20           & 0.01          & 20291        \\ \cline{2-8} 
                               & double     & 50                   & 14         & 14             & 20           & 0.02          & 20291        \\ \hline
\multirow{5}{*}{LSTM}          & copy       & $4 \times 256$       & 128        & 7              & NA           & 0.0002        & 1918222      \\ \cline{2-8} 
                               & reverse    & $4 \times 256$       & 128        & 7              & NA           & 0.0002        & 1918222      \\ \cline{2-8} 
                               & bigramFlip & $4 \times 256$       & 128        & 7              & NA           & 0.0002        & 1918222      \\ \cline{2-8} 
                               & addition   & $4 \times 256$       & 14         & 64             & NA           & 0.0002        & 1918222      \\ \cline{2-8} 
                               & double     & $4 \times 256$       & 14         & 64             & NA           & 0.0002        & 1918222      \\ \bottomrule
\end{tabular}
\end{center}
}
``Vocab'' is the size of the vocabulary (i.e. the total number of possible characters of each input sequence).
``Embed'' is the dimension of the embedding space.
For example, if ``Embed'' is 7, then each character is mapped to a vector in $\R^7$.
``Mem. width'' is the size of each memory vector.
``LR'' is the learning rate.
``\#Param'' gives the total number of trainable parameters.
\label{param_sizes}
\end{minipage}


\textbf{Training and testing.} 
We seek to minimize the negative log likelihood
 of the individual output characters given the input.
All models are trained through RMSProp with momentum .95.
Every epoch has 10 batches, and every batch has 32 instances of the task.
For the LANTM models, after 100 epochs, we half the learning rate if the best error so far is not improved in 30 epochs.
The LSTMs are trained with learning rate 0.0002, with no learning rate adjustments during training.

Since the training sets are large and separate from the test sets, we train until convergence, testing the models periodically --- every 20 epochs for the LANTM models, and every 200 epochs for the LSTM baseline.
After training is complete, the best test scores are tabulated.

We tested the models by drawing 100 batches of random problems and computing fine and coarse scores as in \cite{grefenstette_learning_2015}.
Fine score refers to the percentage of digit or characters (including the EOO marker) that the model correctly outputs.
Coarse score refers to the percentage of total problems that the model answers completely correctly.

{\bf Tweaks to the LANTM model.} We applied two tweaks to the LANTM model: 1) we initialized the mix coefficients for write address and action to strong negative values. This means that the LANTM would tend to write in a straight line. 2) We normalized the step sizes to approximately 1 but did not normalize the initial state step sizes.
We found that these two tweaks improved convergence speed and consistency
\footnote{A video of the read and writes of a LANTM-InvNorm learning the copy task with no biases (tweak 1) is available in the Supplementary Materials.
Compare with the corresponding video with biases.
Details of the videos can be found in appendix \ref{gifs}.
}.
Note that with the second tweak, the ``group'' of actions is no longer a group.
This is akin to restricting the head shifts of an NTM to $+1$ and $-1$ \cite{graves_neural_2014}.

\subsection{python programs}

There are 6 types of programs of integers: addition/subtraction, identity,  multiplication with one small operand, small for loops, variable substitution, and ternary ``$a$ if $b$ else $c$'' statements, as illustrated in table \ref{python_input_output}.

\begin{table}[ht]
\centering
\caption{Example input/output for different types of python programs}
\label{python_input_output}
\tiny {
\begin{tabular}{lll}
\toprule
             & Input & Target \\ \midrule
identity     & print(4103) & 3014.\\
small mult.  & print((14*5608)) & 21587.\\
if then else & print((4242 if 8302$>$6721 else 3716)) & 2424. \\
var. subst.
             & f=3184;print((f-29)) & 33728-        \\
addition     & print((3547+7004)) & 15501.\\
small loop
             & b=1398;for x in range(10):b-=6843;print(b)& 23076-.       \\ \bottomrule
\end{tabular}
}
\end{table}

The models were required to read the input program, which terminates with a ``print'' statement, and output the correct integer response, \emph{in reverse sequence}, without being fed the correct answer (same as in sequence and arithmetic tasks, but different from \cite{zaremba_learning_2014}, which used teacher forcing).
The LANTM models were prohibited from writing during the answer phase, as above.
All input symbols were embedded into $\R^{100}$ before being fed to the machines.

We performed curriculum learning, using the ``mixed'' strategy of \cite{zaremba_learning_2014}, starting from 2 digits operands up to 4 digits operands.
We evaluated the models on their coarse and fine scores on randomly sampled 4 digit programs.
Training was done by RMSProp with learning rate 0.002, which was multiplied by 0.8 whenever the validation accuracy became lower than the highest of the last four.
BPTT was always performed over the entire input and response phase.

The LSTM baseline had a single layer of 128 cells, as did the controllers of LANTM-InvNorm and LANTM-SoftMax, which also had memory width of 128.
This comes out to be 127,890 parameters for the LSTM baseline and 212,149 parameters for the LANTM models.
The LSTM was initialized to have weights uniformly in $[-0.08, 0.08]$ except that the forget gates are set to 1.
The controllers of the LANTM models have weights initialized uniformly in $[-0.0008, 0.0008]$ and the forget gates set to 1 as well.
There were no write biases or normalization of step sizes.

\subsection{language modelling}

The Penn tree-bank corpus consists of 929k/73k/82k train/validation/test words, with a total vocabulary of 10k words.
We followed \cite{zaremba_recurrent_2014} for preprocessing the corpus.
We used batch size of 32.

We embed the words into $\R^{256}$ before feeding into the models.
The LSTM baseline is 1 layer of 128 cells, and the LANTM models have controllers of the same size, along with memory vectors in $\R^{100}$.
This translates to 4,047,632 parameters for LSTM and 4,323,329 parameters for the LANTM models.
The LSTM was initialized to have weights uniformly in $[-0.08, 0.08]$ except that the forget gates are set to 1.
The controllers of the LANTM models have weights initialized uniformly in $[-0.008, 0.008]$ and the forget gates set to 1 as well.
The write biases were set to -10 as in the sequence and arithmetic tasks, but there is no normalization of step sizes.
Whenever the external memories filled up to 100, the oldest 60 memory vectors were discarded.

The number of BPTT steps is 20, and we used Adagrad with learning rate 0.05, which was halved each time the validation perplexity exceeded that of the previous epoch.

\section{Background}
\subsection{Lie groups}
\label{Liegroups}
We here review basic concepts of (Lie) group theory.

A {\bf group} is a set $S$ with operations $*$ (multiplication), $\inv{(.)}$ (inverse), and $e$ (unit) of arity respectively 2, 1, 0, such that
\begin{itemize}
  \item (associativity) for all $a, b, c \in G$, $(a*b)*c = a*(b*c)$
  \item (inverse) for all $a \in G$, $a * \inv a = \inv a * a = e$
  \item (identity) for all $a \in G$, $a * e = e * a = a$
\end{itemize}

The classical examples are $(\Z^n, +, -(.), 0)$, $(\R^n, +, -(.), 0)$, matrix groups like $\GL(n)$, and cyclic groups $\Z/n\Z$.

A group often ``acts on'' another object or set, like a hand twists a rubik's cube.
For example, imagine an equilateral triangle with its vertices colored differently.
Rotating the triangle by 120 degrees permutes the vertex color but leaves the overall shape unchanged.
If we let $0, 1, 2 \in \Z/3\Z$ correspond respectively to rotations of the equilateral triangle by 0, 120, or 240 degrees, and addition in $\Z/3\Z$ corresponds to applying two such rotations consecutively,
then $\Z/3\Z$ is said to act on the set of color permutations of the triangle, because it maps one such permutation to another by a rotation.  
Or, consider $A = \R^2$ as a set of vectors and $B = \R^2$ as a set of points.
One may drag an element of $B$ by a vector from $A$, thus mapping it to another element of $B$.
Then we say $A$ acts on $B$ by vector addition.
As this example illustrates, a group $G$ always acts on itself by the group multiplication (in the example, this is addition of $\R^2$ vectors).
So in fact, every group acts on another set.
Formally, a {\it group action} of group $G$ on set $X$ is defined as a mapping $\phi: G \times X \to X: (g, x) \mapsto g \cdot x$ 
such that
\begin{itemize}
  \item $e \cdot x = x$ for all $x \in X$
  \item $(a*b) \cdot x = a \cdot (b \cdot x)$ for all $a,b \in G, x \in X$.
\end{itemize}

It is the ubiquity of group action that explains the ubiquity of groups in mathematics.
In this paper, we only borrow the language of groups and group actions to the extent it neatly expresses many ideas central to our design.
No advanced ideas from mathematics are used.

A {\it Lie group} is a group with a smooth manifold structure such that multiplication and inverse operations are smooth maps.
Similarly, a {\it smooth group action} of a Lie group $G$ on smooth manifold $M$ is just a group action $\phi: G \times M \to M$ that is smooth.
In the context of smooth Lie group action, we also call elements of $G$ {\it Lie actions}.

The reader who has had no experience with smooth topology need not worry too much about the precise meaning of these definitions beyond the intuition that ``Lie group is a group such that most things you do to it are differentiable'' and ``smooth Lie group action is a differentiable group action''.
Indeed, the only reason we require a Lie group rather than a group is so that its group action yields to gradient descent.
(To that end, it is not strictly necessary for the groups to be infinitely differentiable, but as all common differentiable groups are Lie groups and all groups explored in this paper are Lie group, this distinction is not needed.)
The reader hoping to learn the basics of smooth manifolds and Lie groups can consult John Lee's excellent {\it Introduction to Smooth Manifolds} \cite{lee_introduction_2012}.

\section{Example representation of Lie group actions on the key space}
\label{ExampleReps}
\subsection{Example: The scaling rotation group $\R^* \times \SO(2)$}

The scaling rotation group $\R^* \times \SO(2)$ is the group of linear transformations of $\R^2$ that decomposes into a rotation followed by a dilation (or contraction).

In the specific case of $G = \R^* \times \SO(2)$, the controller would produce 2 numbers $a = \p a t, b = \p b t$, which represents the element
$$v = \begin{pmatrix}
a & -b \\
b & a
\end{pmatrix}
$$
of the group.
The matrix acts on a key $k = (x, y)^T \in \R^2$ by left matrix multiplication

$$v\cdot k = \begin{pmatrix}
a & -b \\
b & a
\end{pmatrix}
\begin{pmatrix}
x \\
y
\end{pmatrix}
$$

This is the same as scaling by the scalar $c = \sqrt{\|a\|^2 + \|b\|^2}$ and then rotating (i.e. left multiplication) by the orthogonal matrix 
$$\begin{pmatrix} 
a/c & -b/c \\
b/c & a/c
\end{pmatrix}$$

Another viewpoint is to treat $(a, b) \in \R^2 - \{0\}$ as the complex number $a + bi \in \C - \{0\}$.
Then one can view the action $v\cdot k$ for $k = (x, y)^T \in \R^2$ as the complex multiplication $(a + bi) (x + yi)$.
\subsection{Example: The rotation group $\SO(2)$}
The rotation, or special orthogonal, group $\SO(2)$ is as its name suggests, the group of all linear transformations of $\R^2$ expressable as a rotation.

When $G = \SO(2)$, we can just modify the scheme from the last example by scaling $(a, b)$ to unit norm, $(\bar a, \bar b) = (a, b)/c$.
The rest will follow just the same.

\section{Videos of read/write}
\label{gifs}

For each task and each of LANTM-InvNorm and LANTM-SoftMax, we created a video of sample read and writes over the course of learning; the entire album is available in the Supplementary Materials.
Each video was created as follows:
\begin{enumerate}
  \item At the end of each epoch, we randomly selected an input of the maximium training length specific to that task (for example, in the case of addition task, two 16-digit numbers interleaved).
  \item We ran the model, with all weights set as trained so far, on this input and record the read and write locations in the key space, along with the strength of each memory vector.
  \item When training is complete, we plot the recording of each epoch in a separate frame, and string them together into a video file.
  		The write locations are marked by red circles, and filled so that a darker fill color means higher memory strength.
  		The read locations are marked by blue disks and connected together by a blue line chronologically (the read line).
\end{enumerate}

Even though we did not explicitly indicate the directionality of the read line, one may infer the directionality of the write sequence by noting that a red circle with white filling marks the beginning of the writes.
Then the read sequence will follow this directionality in all tasks other than the reverse task.

\emph{Analysis.} One sees clearly that LANTM-InvNorm learned to write in a straight line (which is not surprising given our tweaks to the model) and then read along that same line.
On the other hand, LANTM-SoftMax tended to quarantine its read locations to one end of the write line in the reverse, bigramFlip, and addition tasks.
In the copy and double tasks, the read line doesn't stick to the write line as closely with LANTM-Softmax as with LANTM-InvNorm.
This is expected since SoftMax assigns a memory vector with high value just if its location $a$ is closer to the read location $k$ than any other memory vector, whereas InvNorm requires $k$ to be very close to $a$.

\section{Close analysis}
\label{close_anal}
In this section, we discuss the performance of LATNM-InvNorm through various statistics and example input/outputs.
\subsection{Permutation tasks}

%


\subsubsection{copy}
\begin{figure}
\centering
\begin{subfigure}[b]{\textwidth}
\centering
\includegraphics[scale=.6,trim=0 40 0 0,clip]{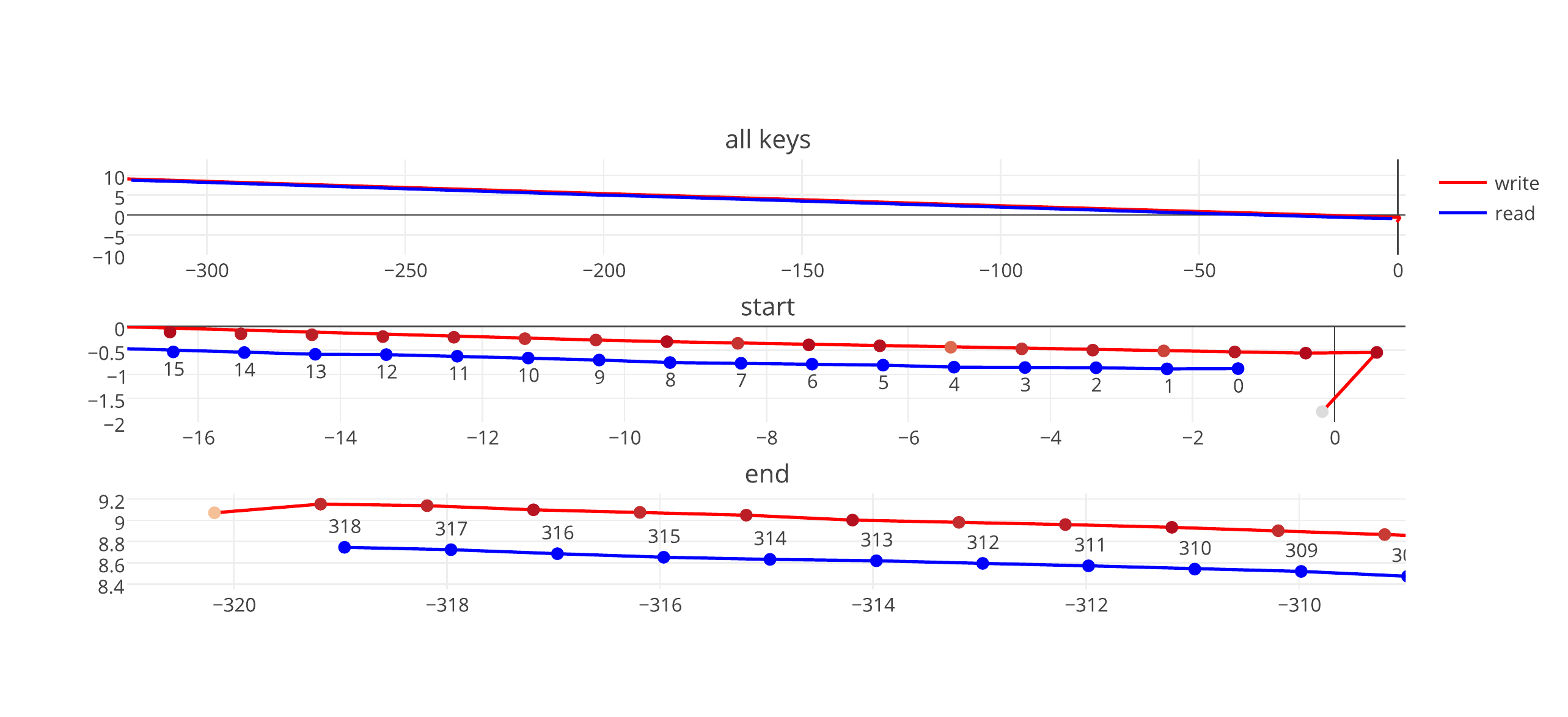}
\caption{\label{copy320keys}}
\end{subfigure}
\begin{subfigure}[b]{\textwidth}
\includegraphics[scale=.6,trim=0 10 0 0,clip]{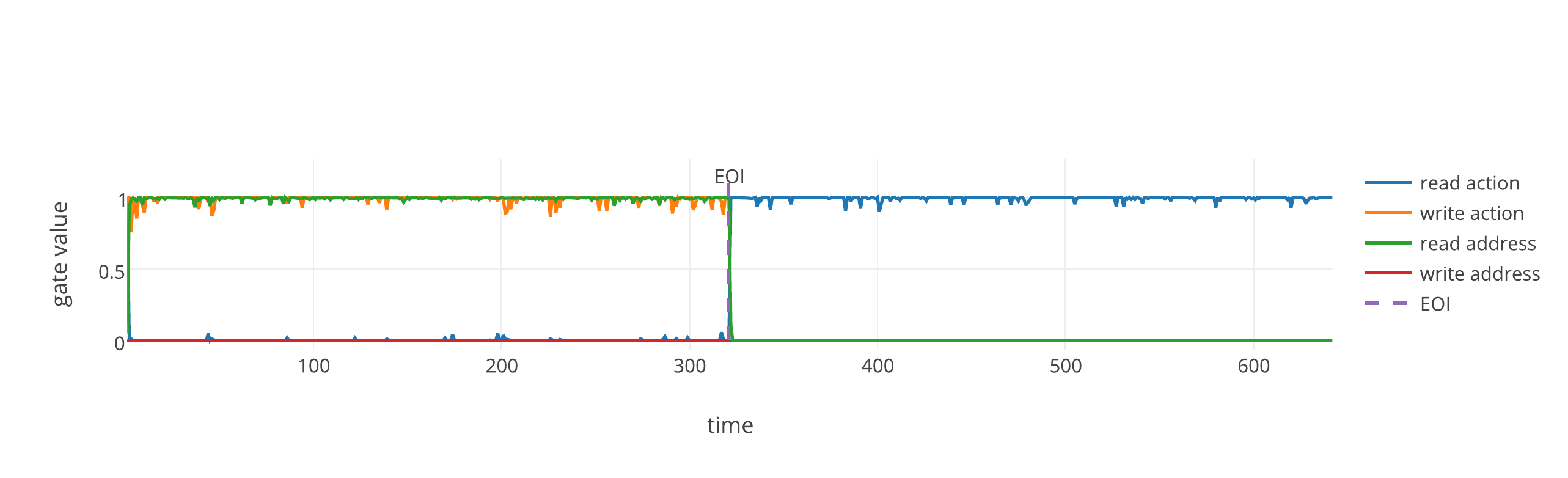}
\caption{\label{copy320gates}}
\end{subfigure}
\begin{subfigure}[b]{\textwidth}
\centering
\includegraphics[scale=.6,trim=0 40 0 0,clip]{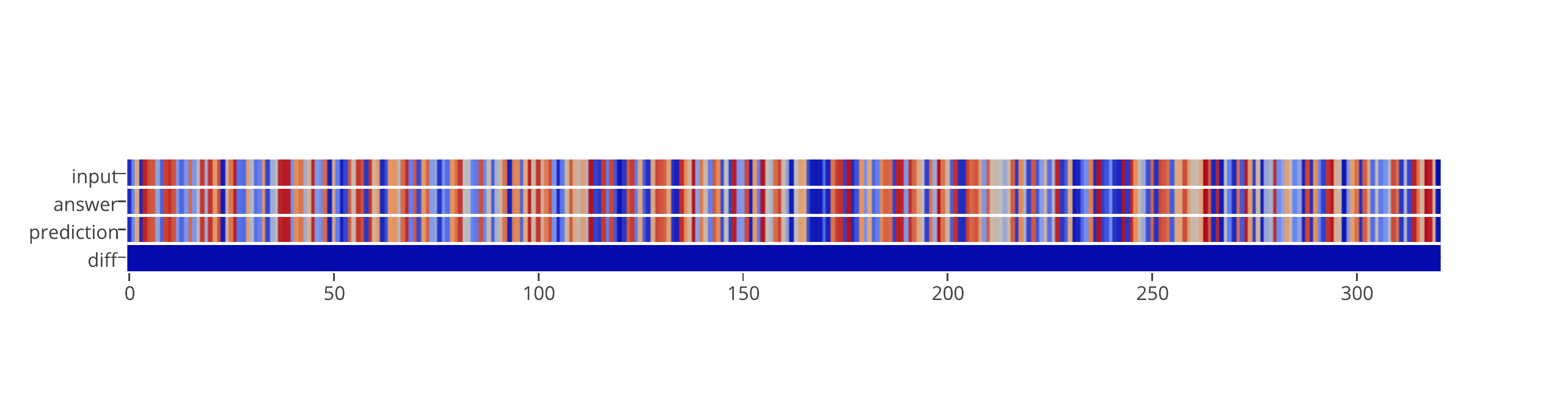}
\caption{\label{copy320anspred}}
\end{subfigure}
\caption{\textbf{Copy task with length 320 input}. Here, a LANTM-InvNorm trained on the copy task of length 1-64 inputs is executed on a length 320 input.
\textbf{(a) read and write locations}.
These are 3 views of the read and write locations of the LANTM.
Red represents write locations, and blue represents read locations.
Only the read locations during the response phase and which are used to compute nonmarker outputs are shown here.
For each red dot, the darker the color the higher the corresponding memory strength.
The scale ratios are all approximately 1:1, i.e. a 1x1 square according to the ticks on the x and y axes should appear as a square.
The subplots, top to bottom, respectively show an overview, the beginning, and the end of the read and write locations in chronological order.
Numbers 0 to 319 label the read keys in this order.
More precisely, read key $i$ is emitted when the LANTM reads the $i$th EOI symbol (so that the actual value read is fed back into the LANTM at time $i+1$).
\textbf{(b) gate values during the execution} The vertical dashed line in the middle (slightly hard to see due to overlap with the green trace) marks the time when the LANTM reads the EOI symbol.
\textbf{(c) correct answer, LANTM's prediction, and the difference}. Each different color represents one of the 128 possible values of the vocabulary. The first row is the correct answer, the second row LANTM's prediction, and the third the difference between the two. In the third row, at most two colors are present: blue means LANTM's response is correct; red means incorrect. In this case there are no red bars because the LANTM was able to perfectly copy the input.}
\label{copy320graphs}
\end{figure}

Figure \ref{copy320keys} shows the read and write locations of such a LANTM-InvNorm, trained on length 1 to 64 input, running on a typical length 320 input.
As one might expect, the reads and writes proceed along straight lines in the key space.
The actual read locations keep close to the corresponding write locations.
In this execution, the LANTM made no errors (figure \ref{copy320anspred}).

Figure \ref{copy320gates} shows the values of the 4 gates governing the computation of read and write keys.
A value of 0 means the gate takes the previous step or key location, while a value of 1 means the gate takes the newly computed step or key location. 
While the write location gates during the input phase and the read location gates during the response phase were as expected pushed to 0, the write step and read step gates were unexpectedly pushed to 1.
Thus the LANTM must have memorized a fixed step size and used it for both reads and writes.

\subsubsection{reverse}
\begin{figure}
\centering
\begin{subfigure}[b]{.37\textwidth}
\centering
\includegraphics[scale=.25,trim=100 40 0 0,clip]{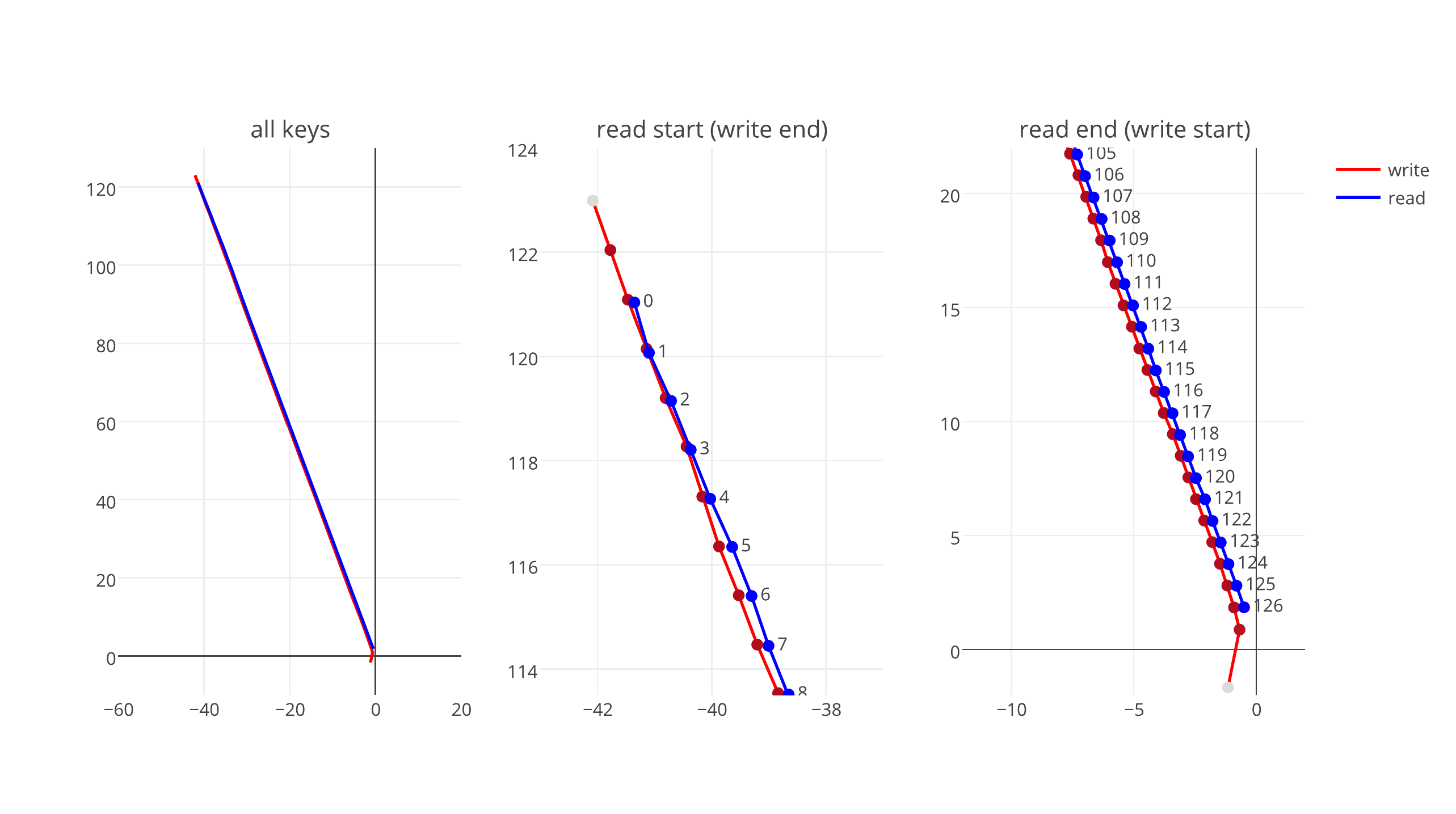}
\caption{\label{reverse218keys}}
\end{subfigure}
\begin{subfigure}[b]{.4\textwidth}
\includegraphics[scale=.3,trim=0 40 0 0,clip]{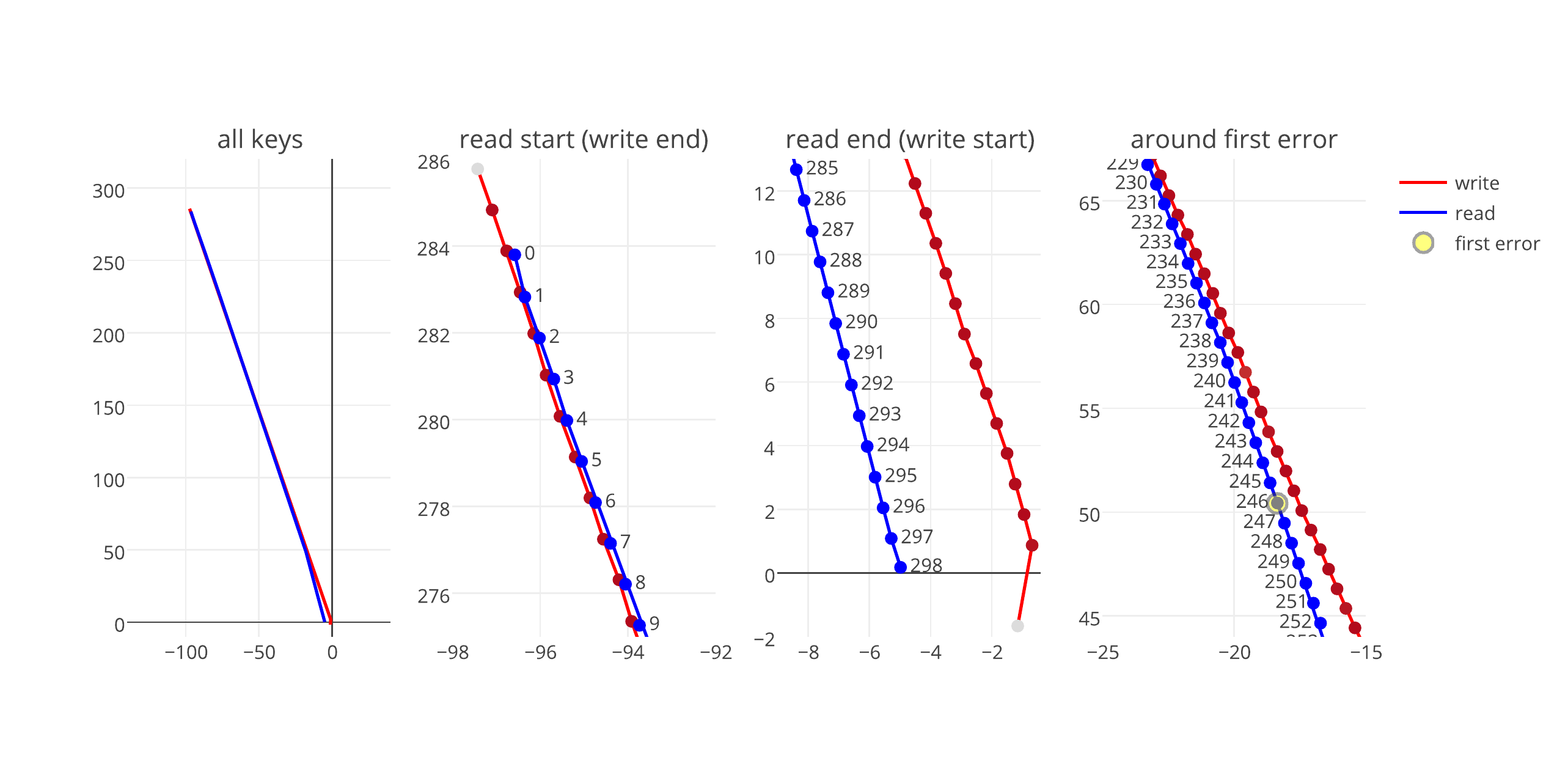}
\caption{\label{reverse300keys}}
\end{subfigure}

\begin{subfigure}[b]{.36\textwidth}
\includegraphics[scale=.25,trim=100 40 0 0,clip]{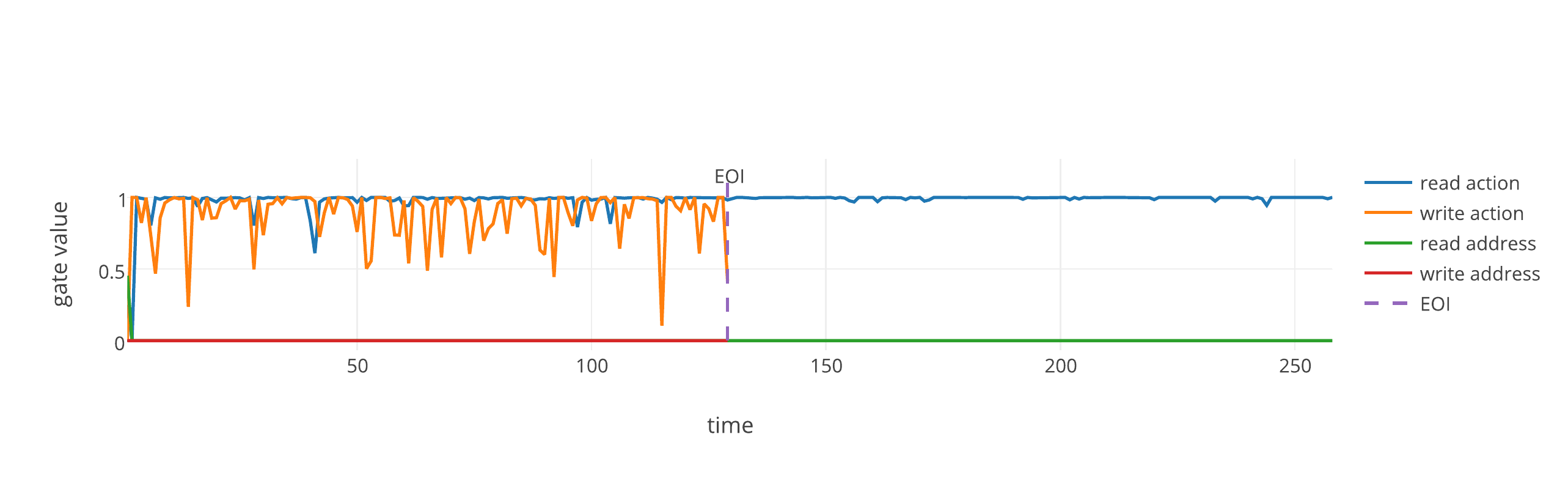}
\caption{\label{reverse128gates}}
\end{subfigure}
\begin{subfigure}[b]{.4\textwidth}
\includegraphics[scale=.25,trim=0 40 0 0,clip]{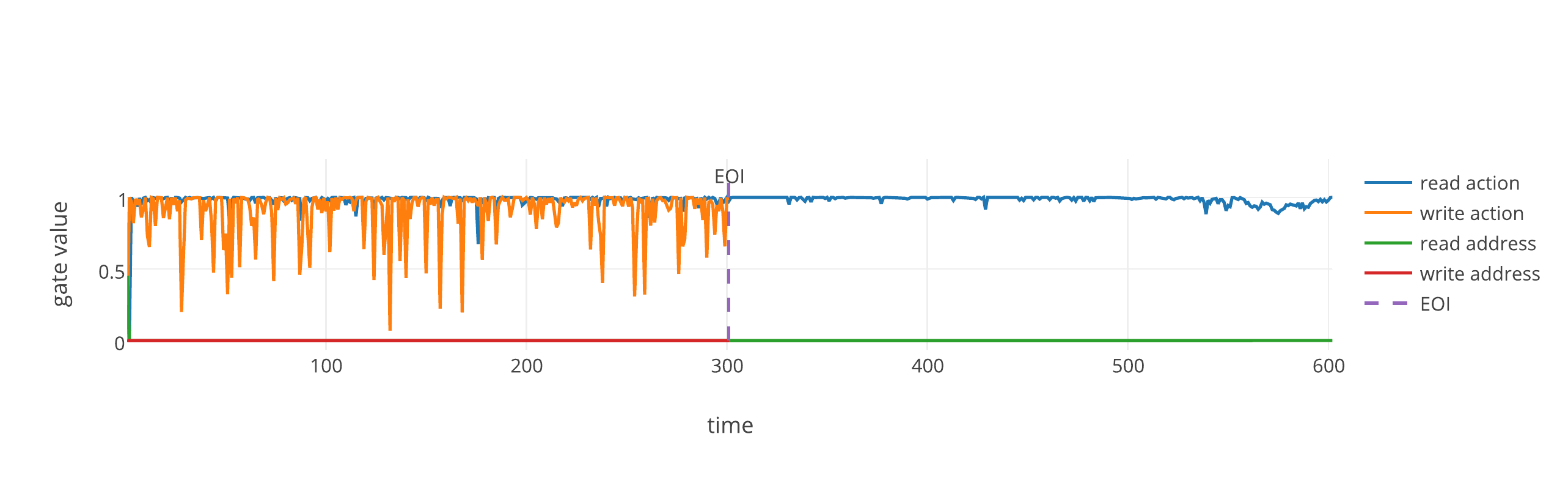}
\caption{\label{reverse300gates}}
\end{subfigure}

\begin{subfigure}[b]{.36\textwidth}
\includegraphics[scale=.25,trim=100 40 0 0,clip]{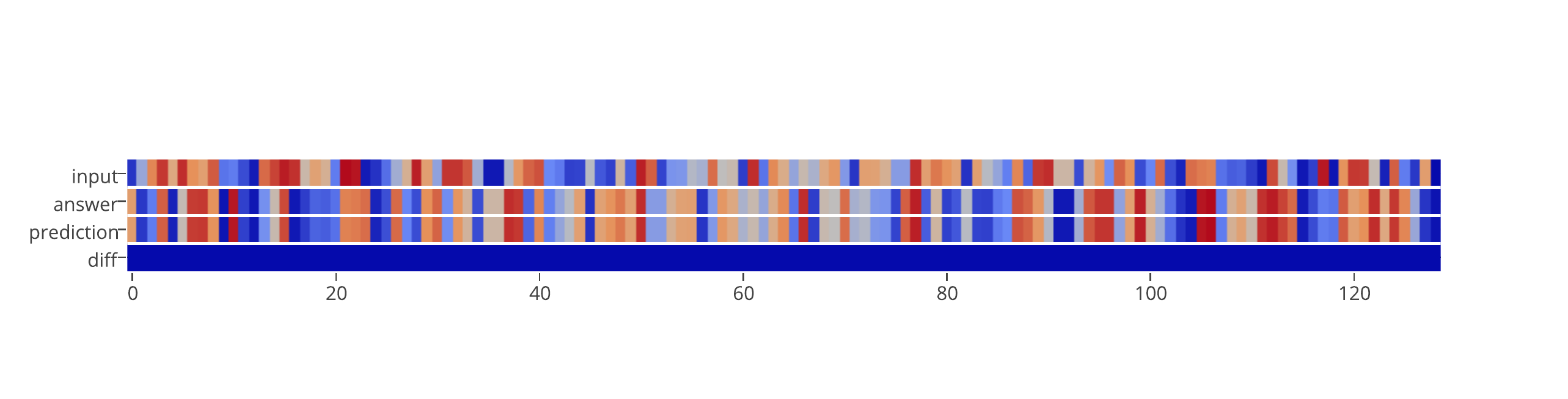}
\caption{\label{reverse128anspred}}
\end{subfigure}
\begin{subfigure}[b]{.4\textwidth}
\includegraphics[scale=.25,trim=0 40 0 0,clip]{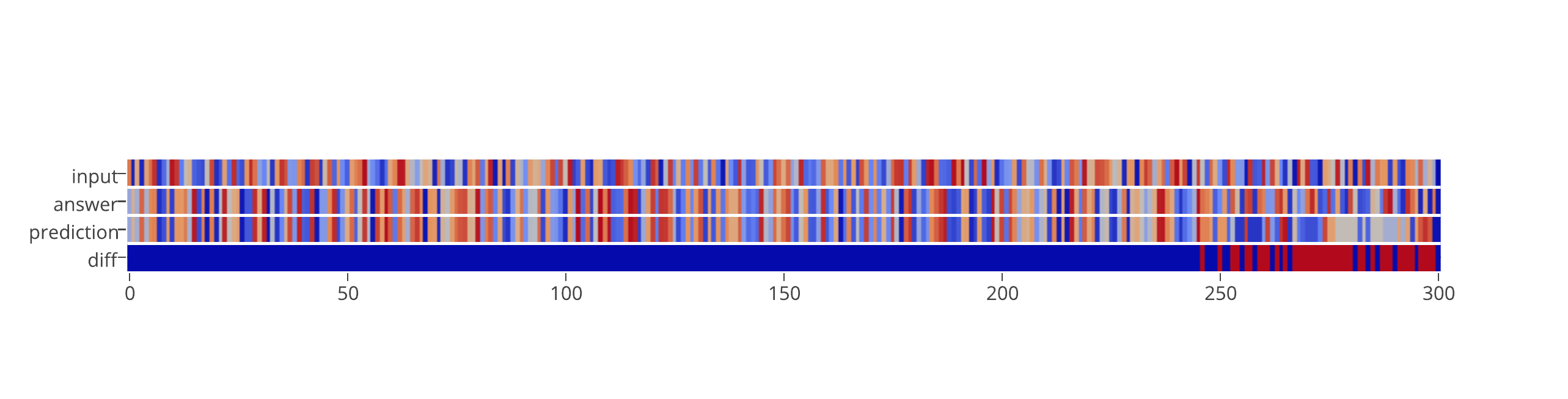}
\caption{\label{reverse300anspred}}
\end{subfigure}
\caption{\textbf{Reverse task with length 128 and 300 inputs.}
The left side shows data on length 128 input; the right side, length 300.
Graph formats are the same as in figure \ref{copy320graphs}, except that in subfigure \ref{reverse300keys}, there is now a fourth plot showing keys around where LANTM made the first mistake. This spot is marked by a yellow dot.}
\label{reversegraphs}
\end{figure}

The counterparts of these graphs for the reverse task are exhibited in figure \ref{reversegraphs}.
On the left we have data for length 128 input, demonstrating a correct execution, while on the right we have data for length 300 input, demonstrating what goes on when extrapolating to higher input sizes.

We see that LANTM trained on the reverse task functions much like that trained on the copy task, with read and write heads traversing on straight lines, except now the directionalities are opposed.
However, when running on length 300 input, the read line, i.e. the curve connecting the read locations in sequence, bends suddenly toward the end, causing almost all reads at the end to diverge from the writes and making almost all outputs at the end to be incorrect. 
This is somewhat surprising, for one might have expected error to come in the form of the accumulation of a small difference between the slopes of the read and write lines.
Along with the sudden dip in read step gate value at the end (blue line in figure \ref{reverse300gates}), the bending of the read line suggests that the LSTM controller started to forget its learned program as the answering phase drew toward a conclusion.

\subsubsection{bigramFlip}
\begin{figure}
\centering
\begin{subfigure}[b]{.37\textwidth}
\centering
\includegraphics[scale=.23,trim=100 40 0 0,clip]{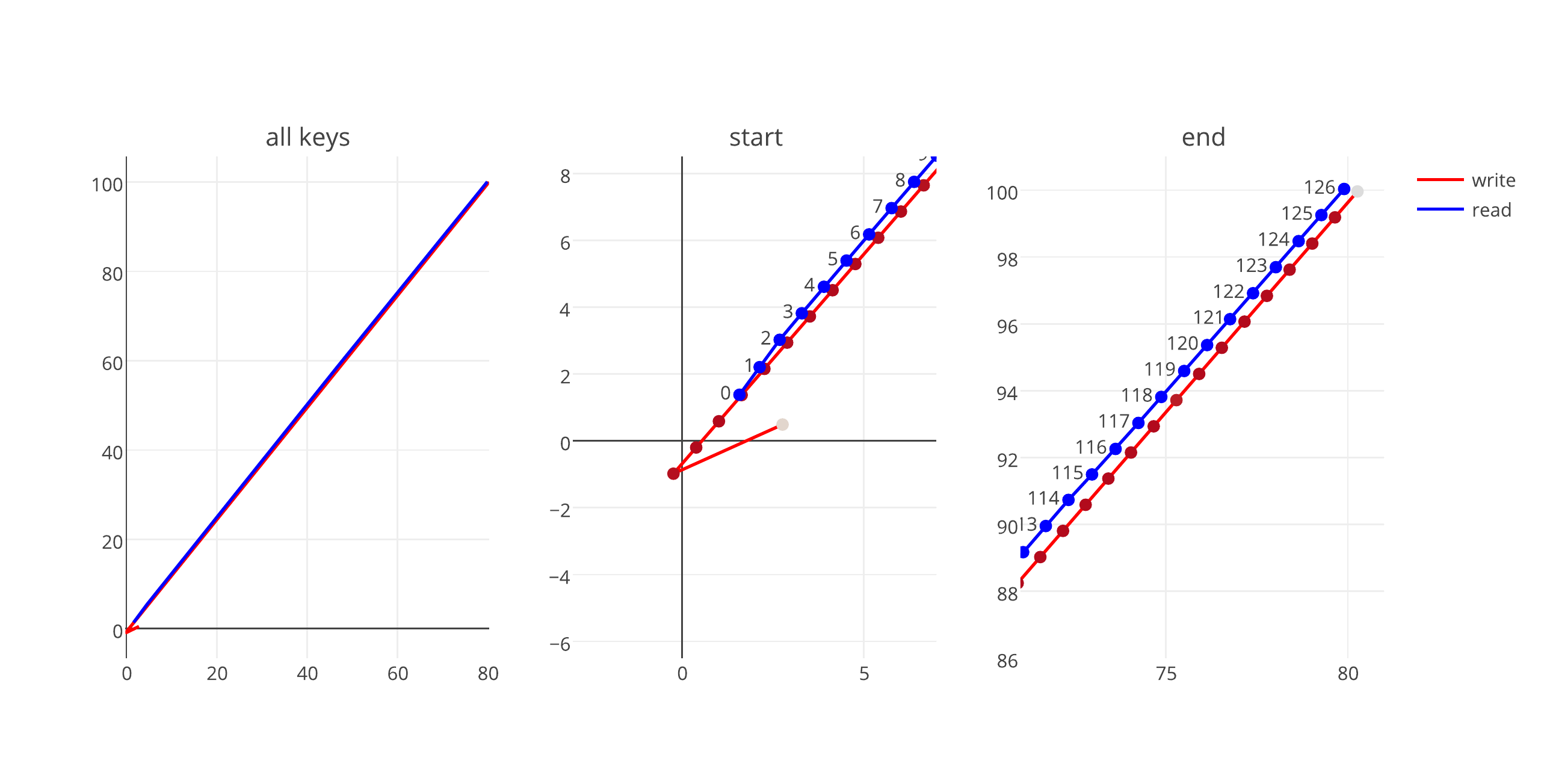}
\caption{\label{bigram128keys}}
\end{subfigure}
\begin{subfigure}[b]{.5\textwidth}
\includegraphics[scale=.23,trim=0 40 0 0,clip]{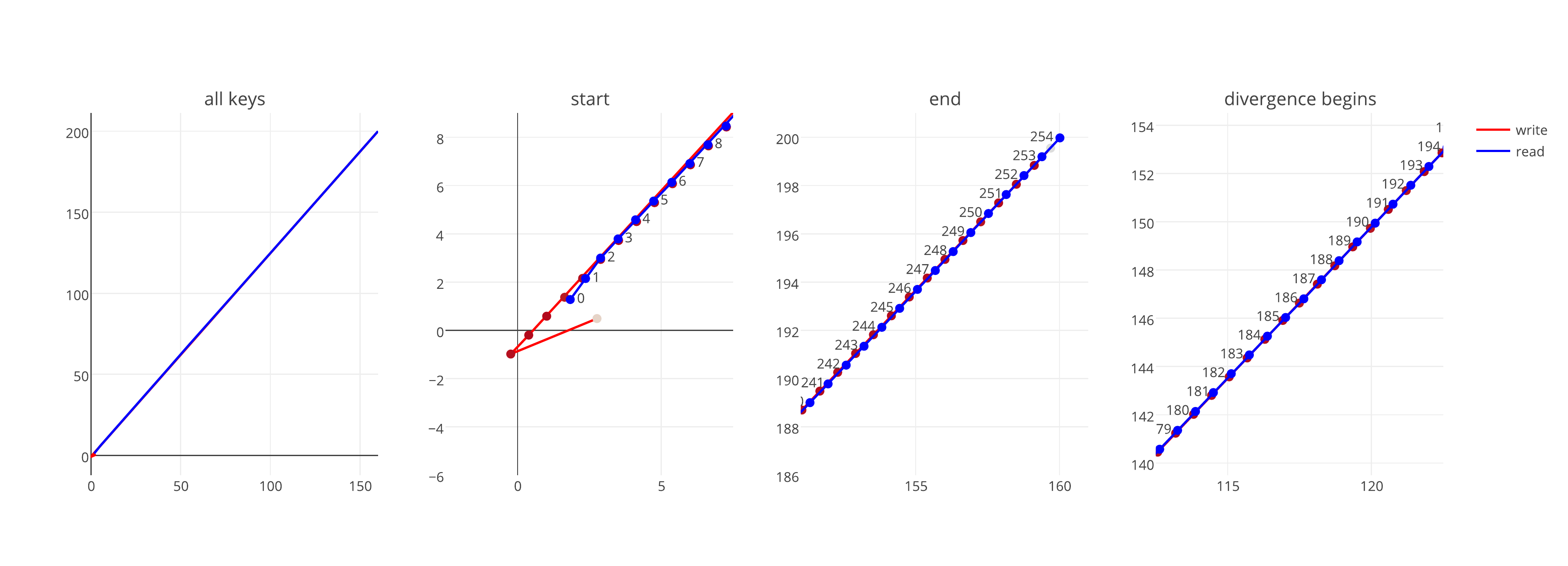}
\caption{\label{bigram256keys}}
\end{subfigure}

\begin{subfigure}[b]{.36\textwidth}
\includegraphics[scale=.25,trim=100 40 0 0,clip]{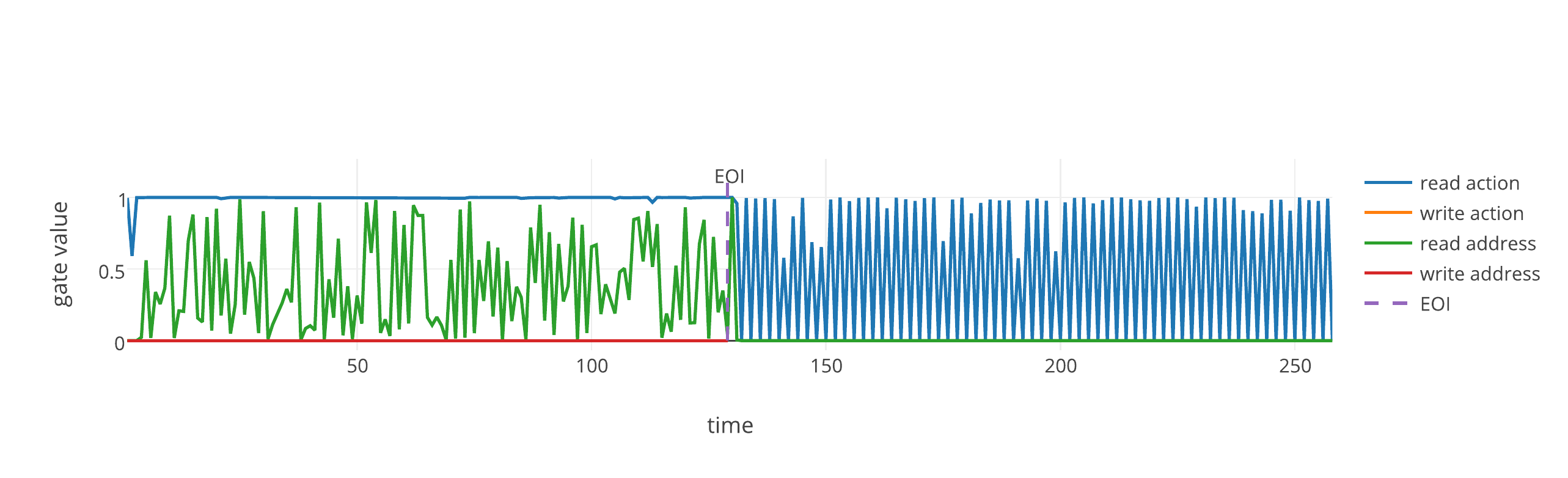}
\caption{\label{bigram128gates}}
\end{subfigure}
\begin{subfigure}[b]{.4\textwidth}
\includegraphics[scale=.25,trim=0 40 0 0,clip]{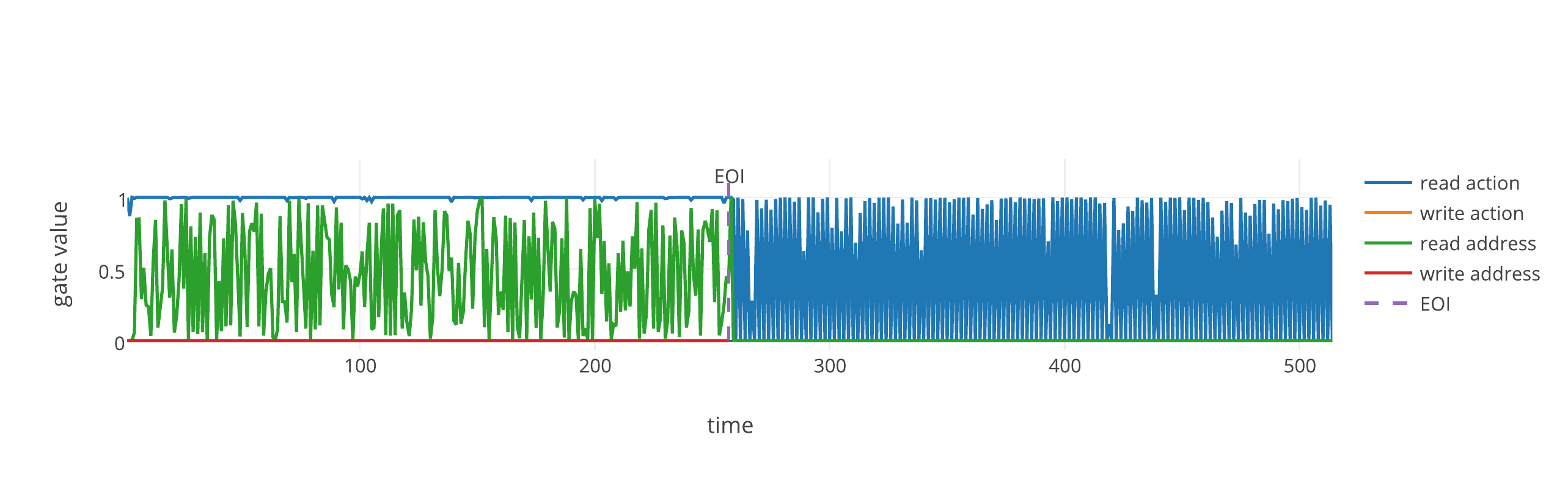}
\caption{\label{bigram256gates}}
\end{subfigure}

\begin{subfigure}[b]{.36\textwidth}
\includegraphics[scale=.25,trim=100 40 0 0,clip]{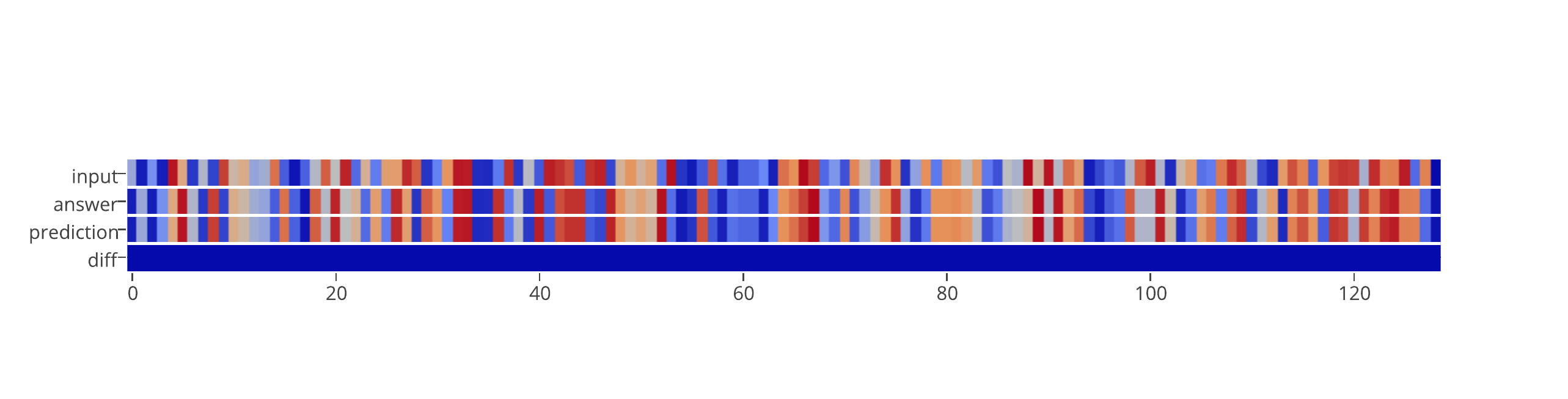}
\caption{\label{bigram128anspred}}
\end{subfigure}
\begin{subfigure}[b]{.4\textwidth}
\includegraphics[scale=.25,trim=0 40 0 0,clip]{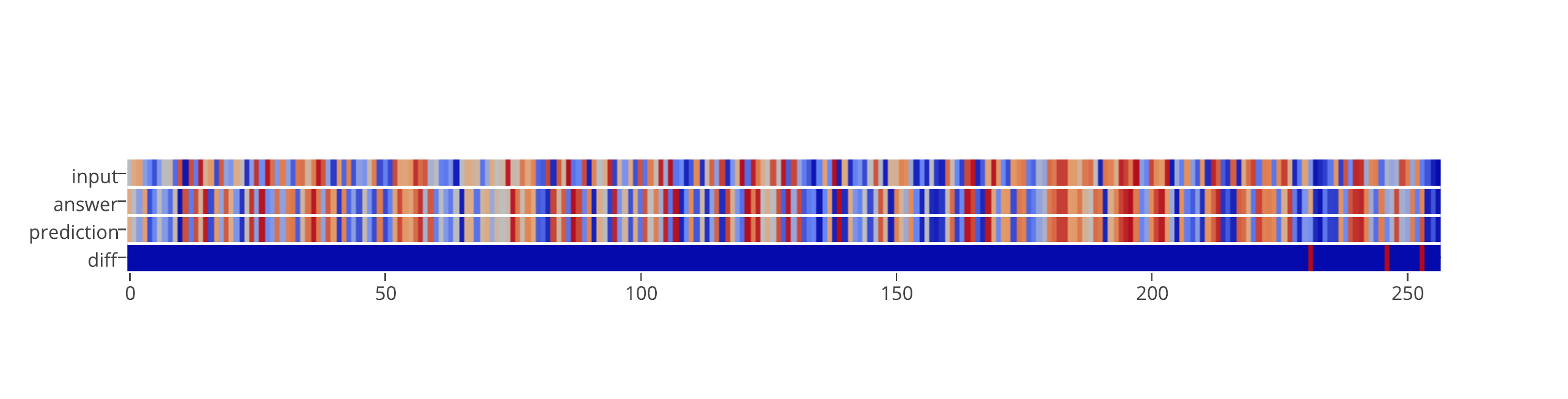}
\caption{\label{bigram256anspred}}
\end{subfigure}
\caption{\textbf{Bigram flip task with length 128 and 256 inputs.}
The left side shows data on length 128 input; the right side, length 256.
Graph formats are the same as in figure \ref{copy320graphs}, except that in subfigure \ref{bigram256keys}, there is now a fourth plot showing keys around where the reads start to diverge from the writes.
In the gate value plots, the write step lines (orange) are almost constant at 0, hidden behind the red lines (write location).}
\label{bigramgraphs}
\end{figure}

The same phenomena appear with the bigramFlip task, where reads and writes happen along 2 closely aligned lines, but when tested by a long input, the reads will abruptly fall out of order: while in the reverse task, the read line visibly bends away from the write line, here the lines stay straight but each step in the read line is elongated, starting around the 187th read (figure \ref{bigram256keys}).

One might be surprised to see that the read happens along a line instead of zigzagging inside the write line.
On closer inspection, we find that LANTM works as follows:

\begin{enumerate} 
  \item LANTM stores representations of the inputs in input order.
  \item Meanwhile it memorizes the first two input characters and outputs them in the reverse order after reading the first two EOI symbols.
  \item When it sees the first EOI symbols, it starts reading the second bigram, i.e. it reads characters 3 and 4 (or their representations in memory; this corresponds to the 5th and 6th memory vectors) after seeing the first and second EOI symbols. This effectively allows it to ``look ahead'' and have each bigram on hand before having to output the flipped image of it.
  \item The LSTM flips each ``look ahead'' bigram and outputs it in order. Repeat for each bigram.
\end{enumerate}

Unique to the LANTM trained on bigramFlip is the oscillation of the read step gate between 0 and 1 (figure \ref{bigram128gates} and \ref{bigram256gates}). This seems like more an artifact of the learning process than a feature of the learned computation, as it would also imply that the controller memorized a single fixed read step, and that the error that occurs with extrapolation seems to stem from the adulteration of this memory.

%
%
%
%
%
%
%
 
\begin{figure}
\centering
\begin{subfigure}[b]{.37\textwidth}
\centering
\includegraphics[scale=.3,trim=100 40 0 0,clip]{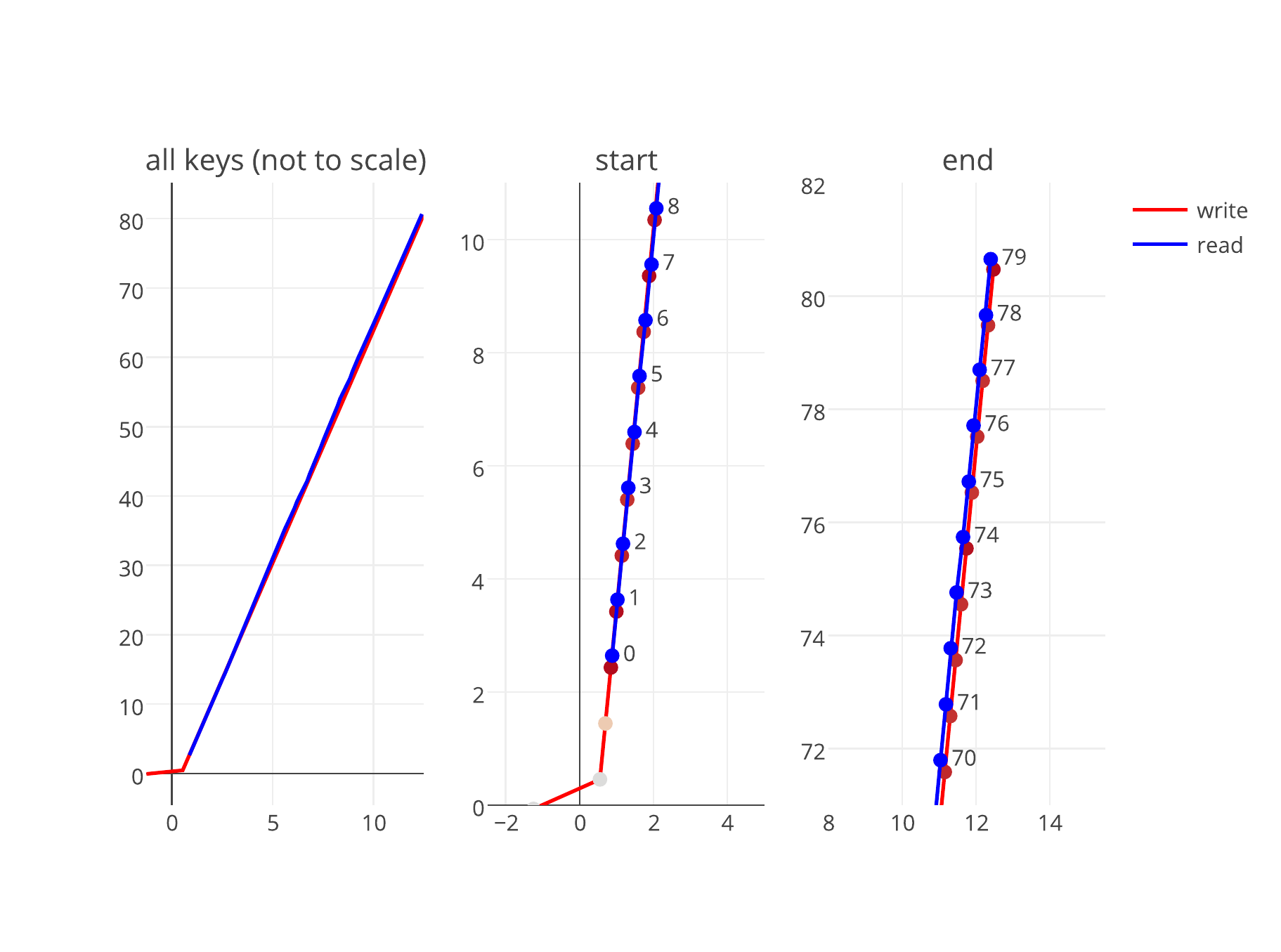}
\caption{\label{double80keys}}
\end{subfigure}
\begin{subfigure}[b]{.4\textwidth}
\includegraphics[scale=.3,trim=0 40 0 0,clip]{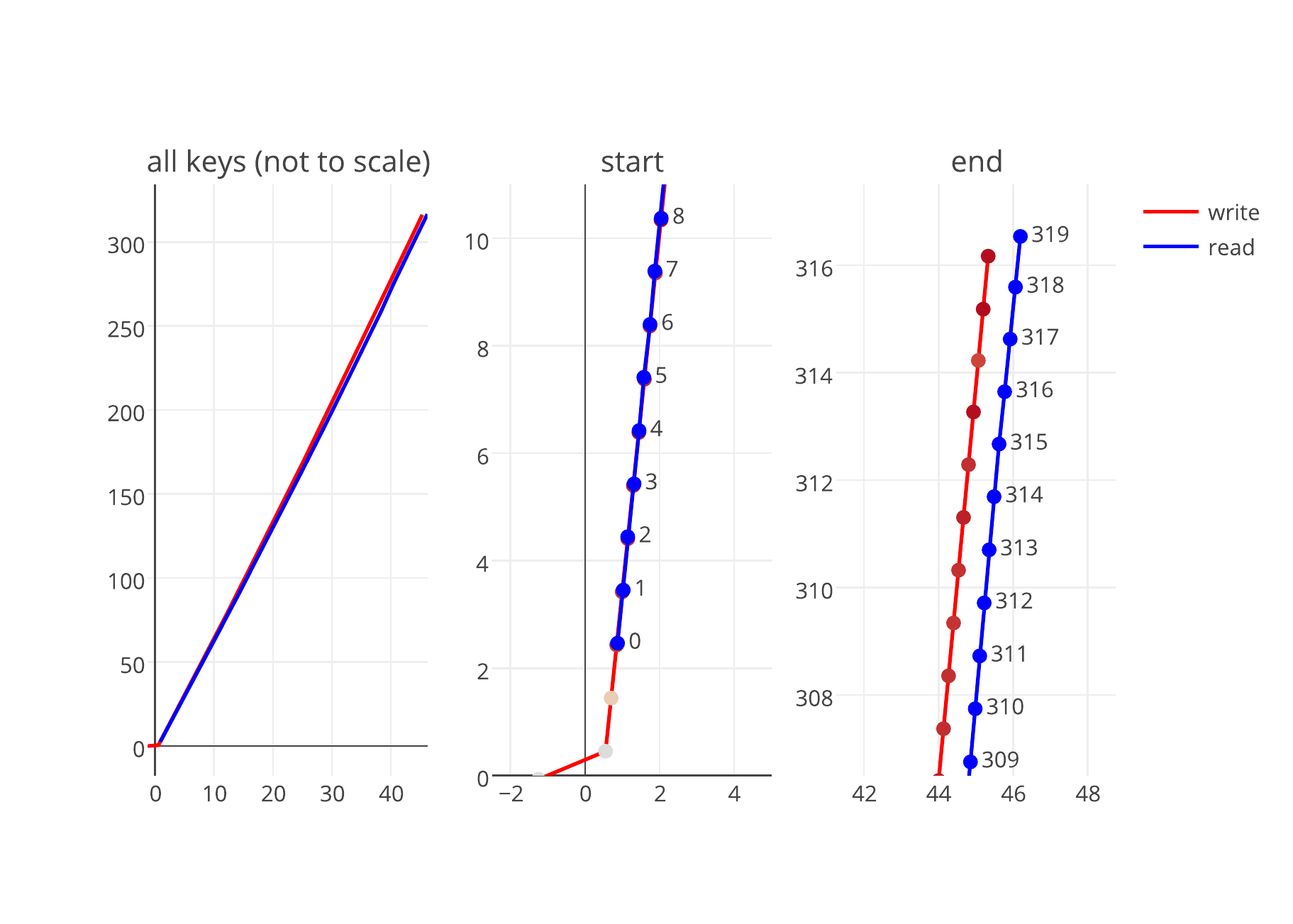}
\caption{\label{double320keys}}
\end{subfigure}

\begin{subfigure}[b]{.36\textwidth}
\includegraphics[scale=.25,trim=100 40 0 0,clip]{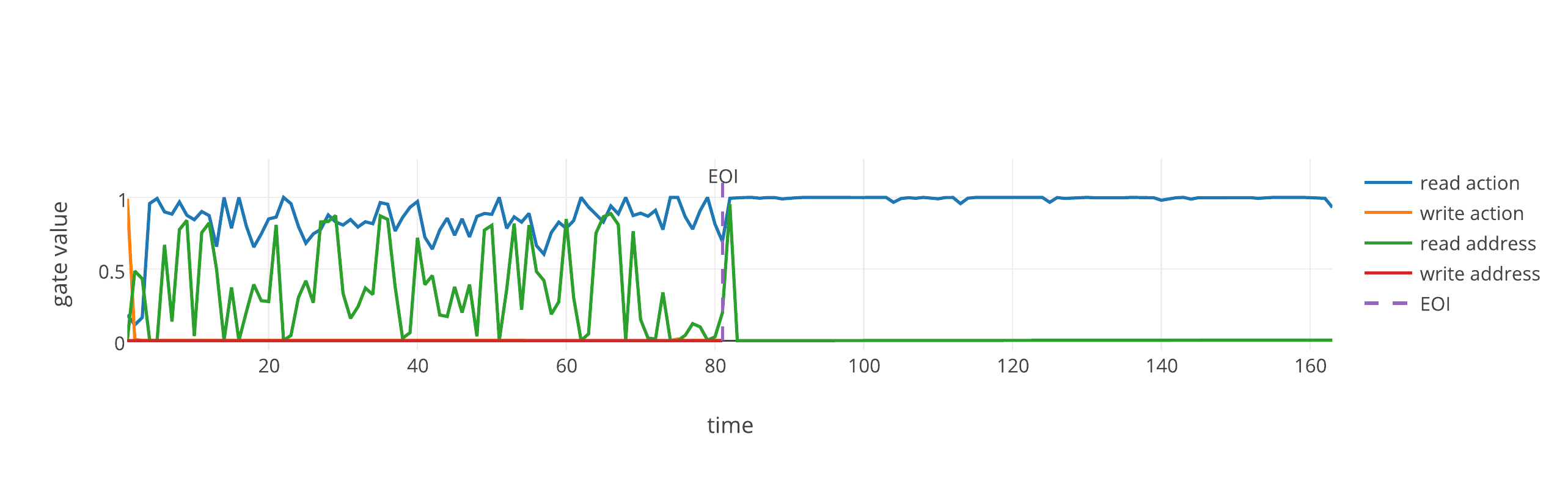}
\caption{\label{double80gates}}
\end{subfigure}
\begin{subfigure}[b]{.4\textwidth}
\includegraphics[scale=.25,trim=0 40 0 0,clip]{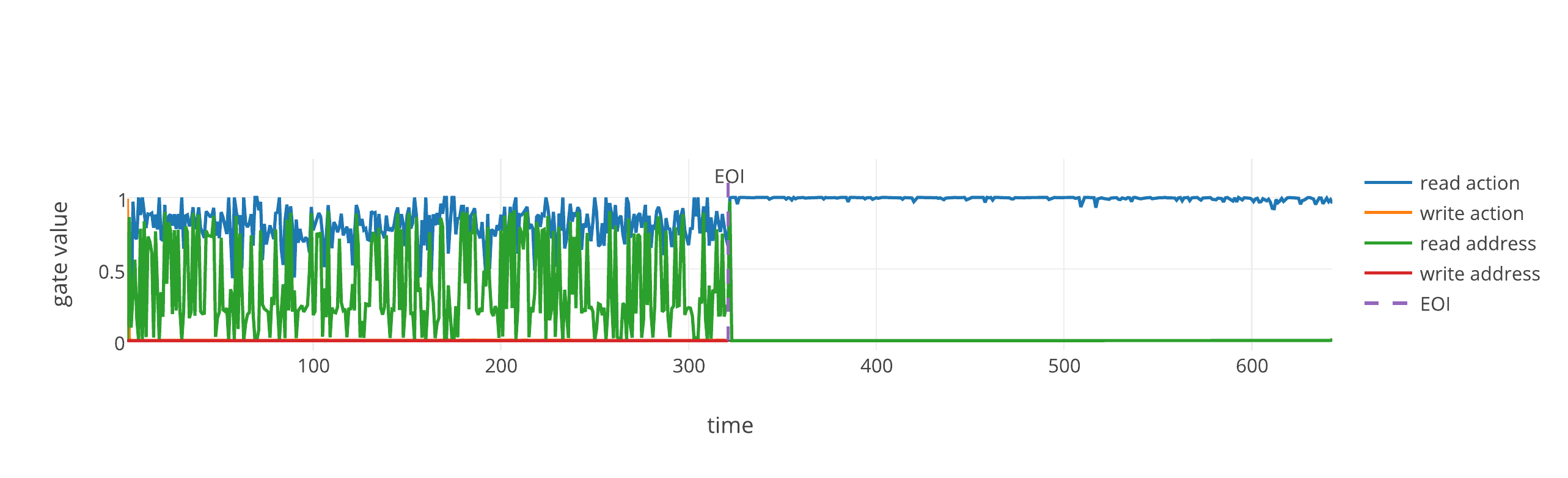}
\caption{\label{double320gates}}
\end{subfigure}

\begin{subfigure}[b]{.65\textwidth}
\tiny
\begin{verbatim}
problem
2 x 
  19405944699084738804681122807647896060125309533161058781560471151464278983184417
=
$038811889398169477609362245615295792120250619066322117563120942302928557966368834
LANTM prediction
$038811889398169477609362245615295792120250619066322117563120942302928557966368834
diff
__________________________________________________________________________________
\end{verbatim}
\caption{\label{double80anspred}}
\end{subfigure}

\begin{subfigure}[b]{1\textwidth}
\tiny
\begin{verbatim}
problem
2 x 
  1168008529246365494001163765851968108385917130977679634159993953145055040607051735753912407298503487088205933788143138139895600272...
=
$02336017058492730988002327531703936216771834261955359268319987906290110081214103471507824814597006974176411867576286276279791200544...
LANTM prediction
$$2233601615442732988800322531703937236677034261955359368319987996290110081214103471507824814597006974176411866576286276279791200544...
diff
_*_*_*******_*___*___*_*__*________*_*_*_**___________*_________*_____________________________________________*_____________________...
\end{verbatim}
\caption{\label{double320anspred}}
\end{subfigure}
\caption{\textbf{Double task with length 80 and 320 inputs.}
The subfigures a, c, e show data on length 80 input; the subfigures b, d, f, length 320.
Graph formats for the first two rows are the same as in figure \ref{copy320graphs}, except that in the first row, the overview plots are magnified horizontally to accentuate the divergence of the read and write lines, demonstrating that the divergence is a gradual build up of difference in slope.
In the gate value plots, the write step lines (orange) are almost constant at 0, hidden behind the red lines (write location).
\textbf{\ref{double80anspred} and \ref{double320anspred}.} We show the doubling problems given to the LANTM and its response.
Dollar signs (\$) represent the end symbol. 
Asterisks (*) mark points where LANTM's answers are incorrect.
In e, only the first (most significant) 130 digits are shown, as there are no errors in the remaining digits.}
\label{bigramgraphs}
\end{figure}

\begin{figure}
\centering
\begin{subfigure}[b]{.4\textwidth}
\centering
\includegraphics[scale=.3,trim=150 40 0 0,clip]{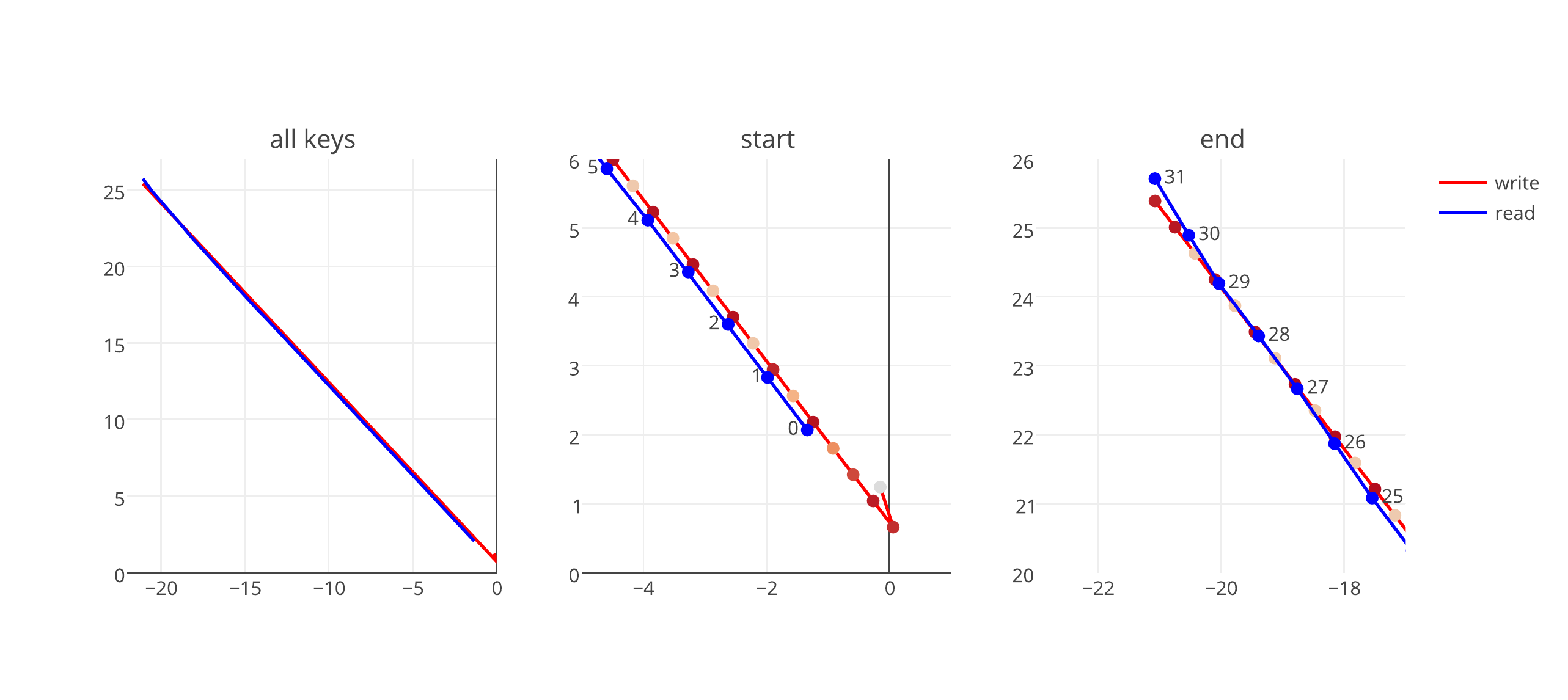}
\caption{\label{addition64keys}}
\end{subfigure}
\begin{subfigure}[b]{.4\textwidth}
\includegraphics[scale=.3,trim=0 40 0 0,clip]{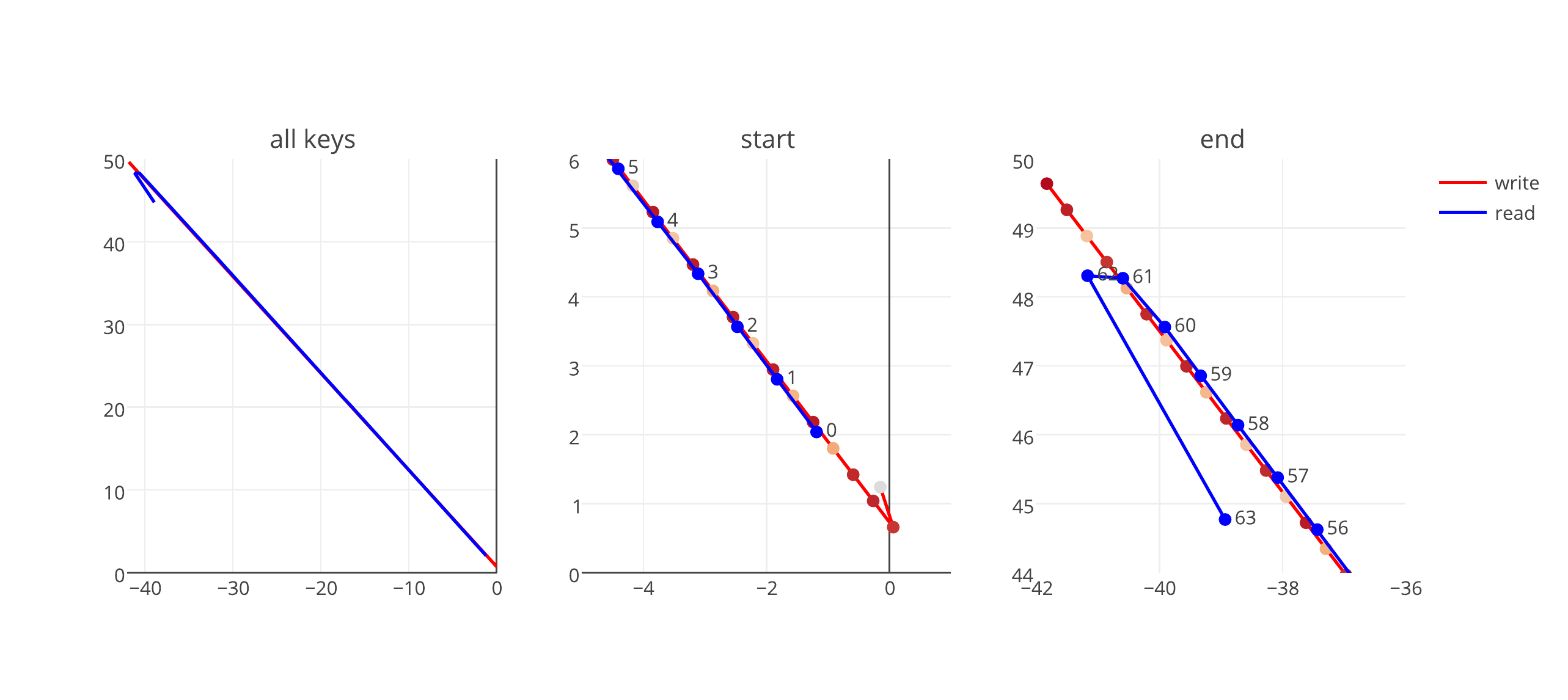}
\caption{\label{addition128keys}}
\end{subfigure}

\begin{subfigure}[b]{.36\textwidth}
\includegraphics[scale=.25,trim=100 40 0 0,clip]{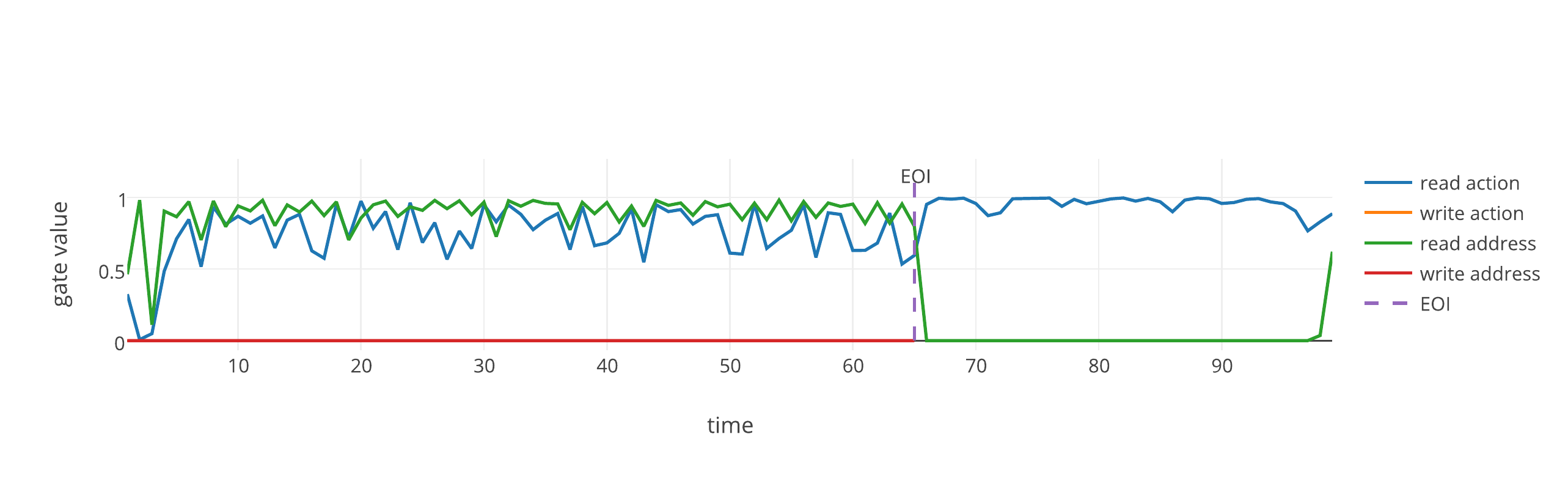}
\caption{\label{addition64gates}}
\end{subfigure}
\begin{subfigure}[b]{.4\textwidth}
\includegraphics[scale=.25,trim=0 40 0 0,clip]{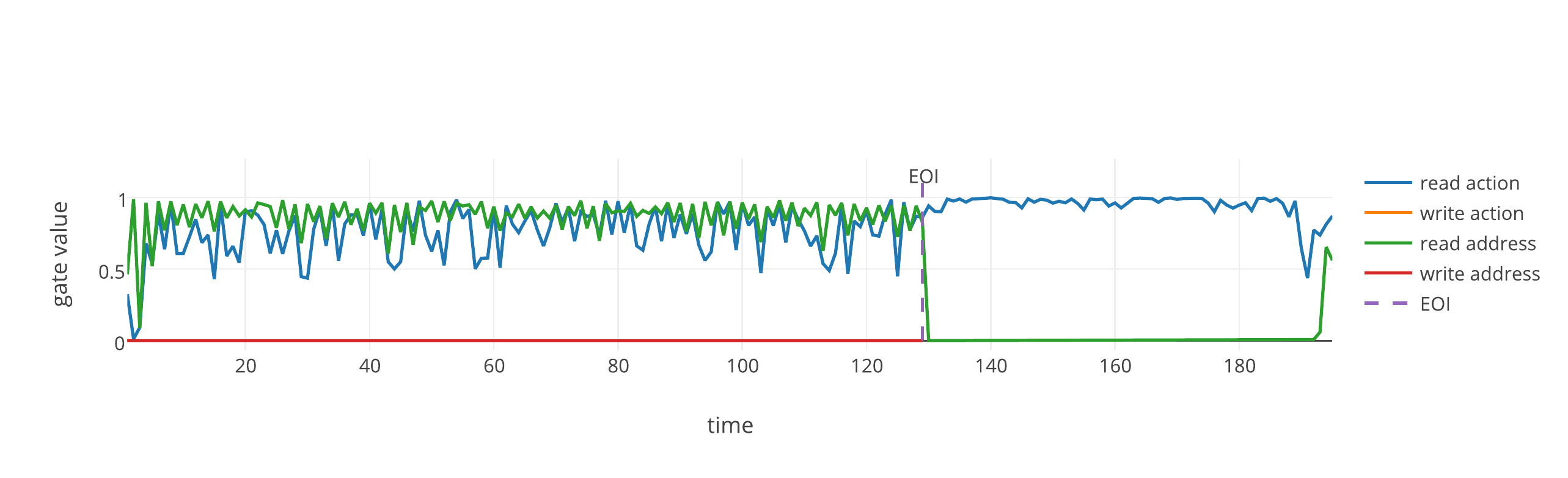}
\caption{\label{addition128gates}}
\end{subfigure}

\begin{subfigure}[b]{.3\textwidth}
\tiny
\begin{verbatim}
problem
  12073190535916485602949518287285
+ 86090378999878149582784619496855
=
$098163569535794635185734137784140
LANTM prediction
$098163569535794635185734137784140
diff
__________________________________
\end{verbatim}
\caption{\label{addition64anspred}}
\end{subfigure}
\begin{subfigure}[b]{.4\textwidth}
\tiny
\begin{verbatim}
problem
  7551725121767466617753259624470709413611531758233121189170215537
+ 6546773181603442580771303866863379302959060107624784946180377555
=
$14098498303370909198524563491334088716570591865857906135350593092
LANTM prediction
$$$$18498303370909198524563491334088716570591865857906135350593092
diff
_****_____________________________________________________________
\end{verbatim}
\caption{\label{addition128anspred}}
\end{subfigure}
\caption{\textbf{Addition task with 32-digit and 64-digit inputs.}
The subfigures a, c, e show data on 32-digit input; the subfigures b, d, f, 64-digit.
Graph formats for the first two rows are the same as in figure \ref{copy320graphs}.
In the gate value plots, the write step lines (orange) are almost constant at 0, hidden behind the red lines (write location).
\textbf{\ref{addition64anspred} and \ref{addition128anspred}.} We show the addition problems given to the LANTM and its responses.
Dollar signs (\$) represent the end symbol.
Asterisks (*) mark points where LANTM's answers are incorrect.}
\label{bigramgraphs}
\end{figure}

\subsection{Arithmetic tasks}

%


In the double task, the LANTM behaved much like it did in the copy task. 
It stored the input in a line and then computed the doubling with carry digitwise.

In the addition task, the LANTM learned to compress each pair of digits of the input numbers (which, as mentioned above, are interleaved) and store them in the odd write locations;
the even write locations had vanishing memory strength (figure \ref{addition64keys} and \ref{addition128keys}).
The LANTM then read off the information by skipping through the odd memory locations.

As with copy and reverse tasks, the read step gate values during the response phase were all close to 1, meaning that the LANTM kept the read step in the LSTM controller memory.
This suggests that the read step gate might be an unnecessary design.

\end{document}